\documentclass[lettersize,journal]{IEEEtran}
\IEEEoverridecommandlockouts                            

\usepackage{graphicx} 
\usepackage{mathptmx} 
\usepackage{amsmath} 
\usepackage{amssymb}  
\usepackage{multirow}
\usepackage{enumerate}
\usepackage{cite}
\usepackage{textcomp}
\usepackage[T1]{fontenc} 
\usepackage[textsize=footnotesize]{todonotes}
\usepackage{xcolor,colortbl}
\usepackage{tabu}
\usepackage{makecell}
\usepackage{array,booktabs}
\usepackage{hyperref}
\usepackage{mathtools, nccmath}
\usepackage{todonotes}
\usepackage{changes}
\definechangesauthor[color=red]{D}
\definechangesauthor[color=red]{IG}
\usepackage{tabularx,booktabs}
\newcolumntype{Y}{>{\centering\arraybackslash}X}

\newcommand*{\figref}[2][]{%
    \hyperref[{#2}]{%
        \ref*{#2}%
        \ifx\\#1\\%
        \else
        #1%
        \fi
    }%
}

\hypersetup{
  colorlinks,
  citecolor=blue,
  linkcolor=blue,
  urlcolor=blue}
  
\usepackage{cite}
\usepackage{textcomp}
\usepackage[T1]{fontenc} 

\begin{document}
	\title{Terrestrial Soft Mobile Robots: A Review
	}
	
\author{Dimuthu~D.~K.~Arachchige%
    \thanks{The author was with the School of Computing, DePaul University, Chicago, IL, USA, and is now with the Department of Computer Science, Hampton University, Hampton, VA, USA.\\
    {\ttfamily\small dimuthu.arachchige@hamptonu.edu}\\[2mm]
    This manuscript is adapted from Chapter~2 of the author's 2024 Ph.D. dissertation, titled ``Modular Approach to Soft Mobile Robots,'' 
    \href{https://via.library.depaul.edu/cdm_etd/60/}{via.library.depaul.edu/cdm\_etd/60/}, 
    and has been reviewed only by the dissertation committee.}
}

	\maketitle

	\begin{abstract}
        Soft mobile robots have emerged as a promising area of research with potential applications in various disciplines including but not limited to search-and-rescue, service, surveillance, explorations, and manufacturing. In this article, we provide a comprehensive review of the current state of soft mobile robot research, focusing on wheelless terrestrial locomotive systems. 
        We include past and present developments in locomotion strategies, actuation methods, modeling approaches, and control systems. Further, we identify key research challenges that must be overcome to enable the widespread adoption of soft mobile robots in various applications. Overall, this article provides a valuable resource for researchers and practitioners interested in the field of soft mobile robots and soft robotics.
	\end{abstract}


	\IEEEpeerreviewmaketitle
  
  \section{Introduction\label{sec:Introduction}} 	

    Soft robots are designed using pliable materials such as silicone or rubber, setting them apart from traditional robots constructed from rigid components like metal or plastic \cite{rus2015design}. This distinct composition allows soft robots to adapt and mold to their surroundings, positioning them as ideal candidates for tasks where conventional robots may falter. Soft Robotics emerges as an interdisciplinary field, intersecting with material science, actuator and sensor technology, design and fabrication techniques, control and modeling strategies, along with studies in locomotion, manipulation, bioinspiration, and biomechanics \cite{whitesides2018soft}. Within this broad domain, soft robotic locomotion represents a specialized area of study focused on the innovative movement and navigation capabilities of soft robots \cite{calisti2017fundamentals}. Researchers are actively developing diverse soft robotic locomotion systems, seeking novel design and control approaches to enhance these robots' mobility. This endeavor aims to transition soft robots from the confines of laboratory research to practical, real-world applications \cite{mazzolai2022roadmap,chen2020soft,ahmed2022decade}. The advent of soft robotic locomotion holds the promise of transforming the robotics field by enabling the execution of tasks previously deemed challenging or unattainable for standard mobile robots.

\subsection{Motivation for Soft Mobile Robots}\label{subsub:Motivation}

Soft mobile robots have a wide range of potential applications due to their unique capabilities, including flexibility, adaptability, safety, and the ability to move through challenging environments. Some of the most promising applications of soft mobile robots include; exploration, agriculture, and manufacturing \cite{el2020soft}. They can be used to explore challenging and hazardous environments, such as space, deep oceans, and disaster zones, where traditional robots and humans may face difficulties or danger \cite{aracri2021soft}. For example, soft mobile robots could be designed to crawl through narrow passages, squeeze through small openings, and navigate uneven terrain \cite{hawkes2017soft}. Soft locomotive systems that have a small cross-section-to-length ratio such as soft robotic snakes, soft eel robots, etc., could be used to inspect and repair pipelines or other infrastructure in hazards (e.g. radioactive environments) or hard-to-reach locations \cite{oshiro2018soft}. 

 \begin{figure}[tb] 
		\centering
		\includegraphics[width=1\linewidth]{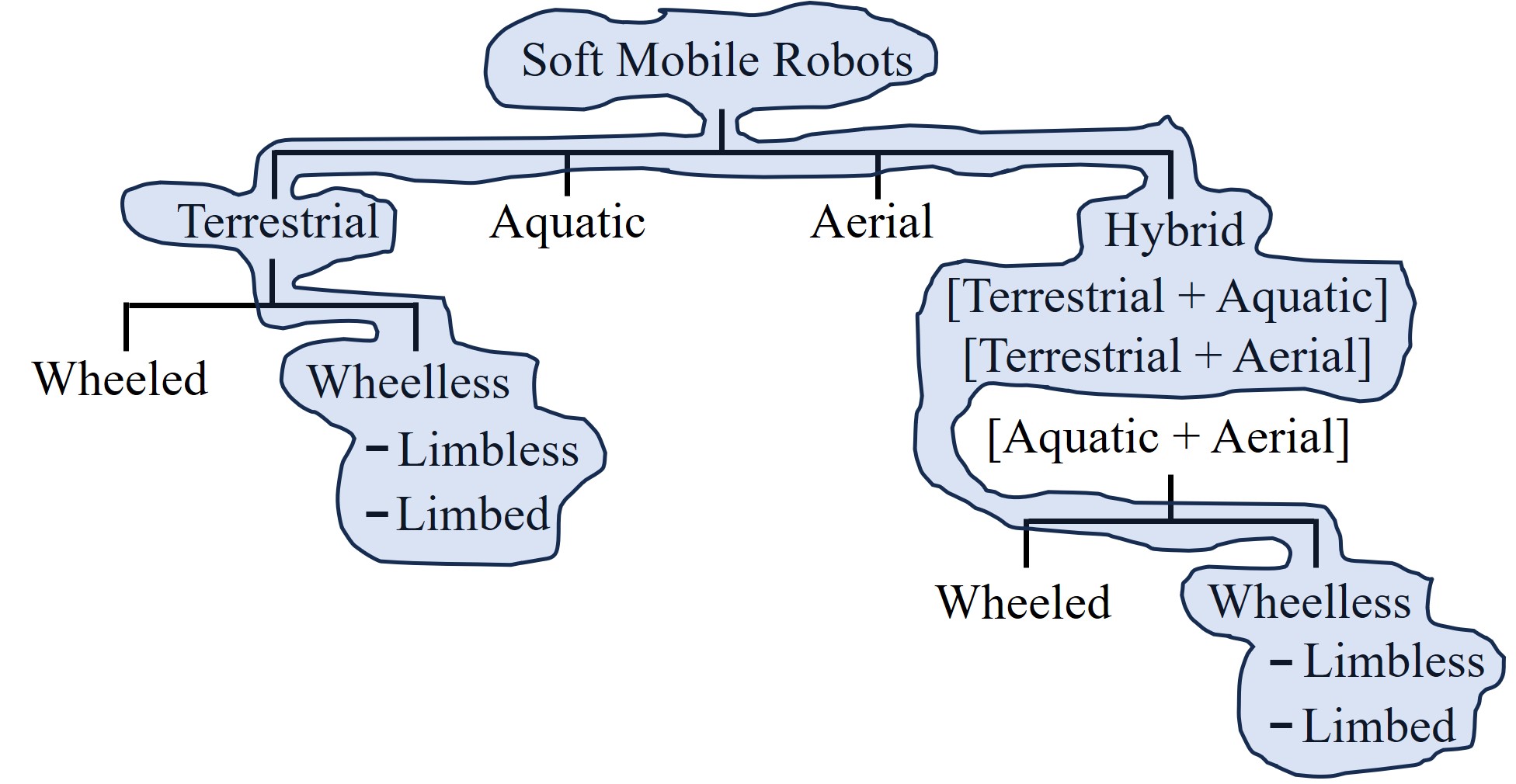}
		\caption{Classification of soft mobile robots highlighting the scope of this review.}
		\label{fig:FIg1_SoftMobileRobotCategory} 
	\end{figure}

 \begin{figure*}[tb] 
		\centering
		\includegraphics[width=0.9\linewidth]{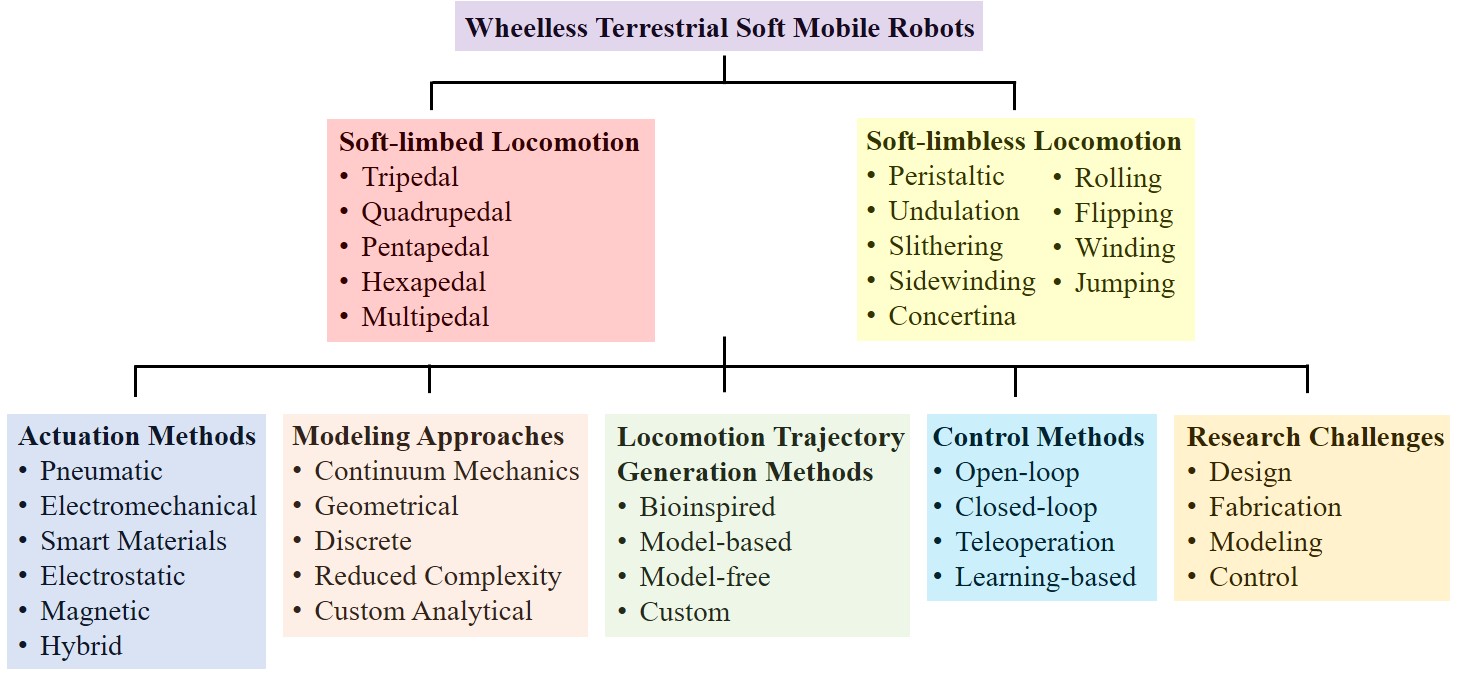}
		\caption{Summary of the topics discussed in this review.}
		\label{fig:Fig2_ReviewOverview} 
\end{figure*}

Furthermore, soft locomotive systems can be used in a variety of ways in planetary exploration \cite{ng2021untethered}. Traditional robotic exploration systems, such as rovers, face limitations in their ability to traverse rough terrain and often require complex and expensive control systems. Soft robots, on the other hand, can use their deformable and adaptable bodies to move through environments that are difficult for traditional robots to navigate, such as caves, steep slopes, and rocky terrain \cite{zhang2022progress}. Additionally, they can be designed to be lightweight and portable, making them easier and cheaper to transport to remote locations \cite{schmitt2018soft}. They can also be used to collect samples or perform other tasks that require a gentle touch, such as studying fragile ecosystems or conducting archaeological excavations.

Soft mobile robots have the potential to revolutionize farming practices by performing tasks such as harvesting, pruning, and plant inspection \cite{kondoyanni2022bio}. Their soft and delicate touch can help avoid damaging crops and reduce waste, and their adaptability allows them to adjust to different crop types and sizes \cite{gunderman2022tendon}. More importantly, they can be used in manufacturing settings to perform tasks such as assembly, packaging, and quality control. Their flexibility and adaptability make them safe collaborative robots that can be employed with humans \cite{ye2022soft}. Therein, they are ideal for handling delicate and irregularly shaped objects, reducing the need for specialized machinery. Over the years, researchers have developed various soft mobile robot prototypes and tried to harness this immense potential \cite{sun2021soft}.


  \subsection{Types of Soft Mobile Robots}\label{subsub:TypesofSoftMobileRobots}

Soft mobile robots can be classified into three main categories based on their operating environments: terrestrial robots, designed for movement on land; aquatic robots, tailored for underwater environments; and aerial robots, engineered for navigating through the air. Additionally, there are robots capable of navigating in hybrid environments, such as those encompassing land, water, and air. Terrestrial robots can be further subcategorized into wheeled systems -- which use wheels as their primary means of locomotion and wheeless systems -- which use various locomotion strategies, such as legged locomotion, snake-like motion, or track-based locomotion. Wheeled terrestrial robots are not suitable for all types of terrain, which has led to the development of wheeless terrestrial robots. These robots are designed for movement on rough or uneven terrain, such as sand, rocks, or steep slopes, where traditional wheeled robots would struggle to operate.

  Wheeless systems can be further classified as sof-limbed and soft-limbless (or soft-bodied) robots. Soft-limbed robots are designed to move using legs. Soft-limbless robots are designed without traditional limbs or appendages, instead relying on their soft and compliant bodies to move and manipulate their environment. Figure \ref{fig:FIg1_SoftMobileRobotCategory} shows an overview of the aforementioned classification.

    \subsection{Scope of the Review}\label{subsub:Scope}
    
    In this study, we focus on wheelless soft mobile robots that exhibit terrestrial locomotion, as illustrated in Fig.~\ref{fig:Fig2_ReviewOverview}. Consequently, our review excludes aquatic, aerial, and hybrid systems lacking the capability for terrestrial movement. We explore robots with various terrestrial locomotion strategies, including jumping and climbing gaits -- encompassing branch climbing, tree climbing, rod climbing, and wall climbing, among others. Additionally, we delve into amphibian-inspired robots that demonstrate terrestrial movement, drawing inspiration from creatures such as sea stars (or starfish), walruses, seals (pinnipeds), and others. Soft rolling robots are also covered in this review. Hence, the "wheelless" is referred to as the one that does not have passive or active wheels. In addition to that, robots that locomote in-body environments such as human or animal organs are not discussed here. 

	Building on the structure presented in Fig.~\ref{fig:Fig2_ReviewOverview}, this review is organized into nine sections. Sec.~II and Sec.~III categorize the fundamental locomotion methods of wheelless terrestrial soft-limbed and soft-limbless robots, respectively. The actuation technologies of such robots are presented in Sec.~IV. Sec.~V explores the different modeling approaches and their unique features thereof. The trajectory generation methods are presented in Sec.~VI, while the locomotion control methods are examined in Sec.~VII. Finally, research challenges and conclusions are summarized in the last two sections.

    \begin{figure*}[tb] 
		\centering
		\includegraphics[width=1\linewidth]{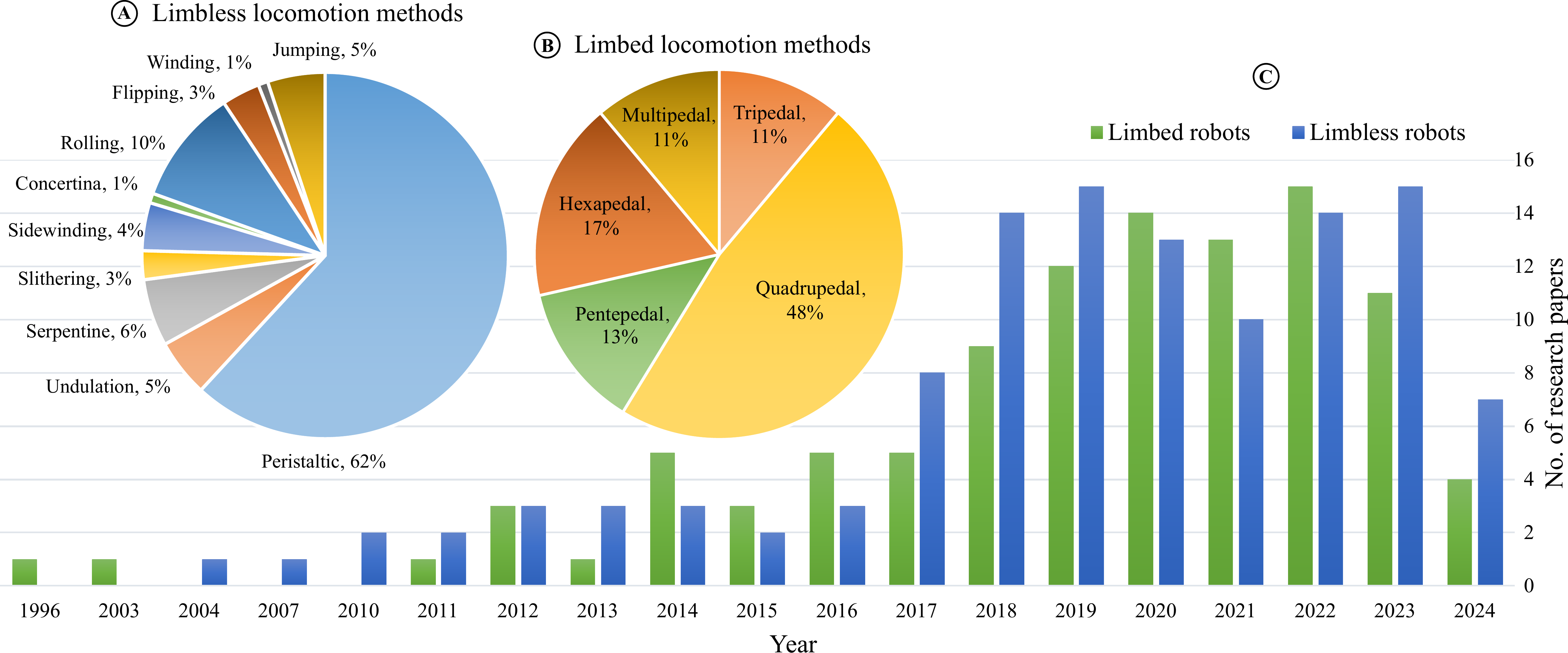}
		\caption{Distribution of research papers on soft mobile robots from 1996 to 2024, as considered in this review article.}
		\label{fig:Fig3_NumberofPapers} 
	\end{figure*}

    \section{Soft-Limbed Robot Locomotion}\label{sec:Soft-LimbedRobotLocomotion}

    Soft-limbed robots represent the most favored subset within the domain of soft mobile robotics, primarily due to their potential for tackling challenging locomotive tasks. Their classification into distinct topologies depends on the number and arrangement of limbs, encompassing categories like tripod, quadruped, pentapod, hexapod, and multipod robots. Depending on these specific topologies, these robots have embraced a variety of locomotion methods, which are discussed in the following sections. Figures~\figref[B]{fig:Fig3_NumberofPapers} and \figref[C]{fig:Fig3_NumberofPapers} show the distribution of soft-limbed robots over the years.

    \subsection{Tripedal Locomotion} \label{subsec:Tripods}

This is a type of locomotion that involves three legs instead of the typical four or six found in most robots. This is inspired by the way animals like spiders, crabs, and tripod fish move, and it has some advantages over other forms of locomotion in certain situations. A few tripodal soft robot prototypes have been proposed to date. Authors in \cite{mao2016locomotion} proposed several multi-limbed 3D-printed soft robot prototypes. One of them is a  tripodal robot that has tri-symmetrical limbs. It operates based on the crawling gaits of a three-leg starfish. The tripod reported in \cite{bern2019trajectory} is a simple foam robot that has three serially arranged straight legs. Therein, the outer two legs are unactuated and provide stability while its central leg contracts and move the robot. The work reported in \cite{wang2021design,perera2023teleoperation,arachchige2023study, arachchige2024tumbling} proposed modular soft robots that have four identical limbs arranged in a spatially symmetric tetrahedral topology (Fig.~\figref[A]{fig:Fig4_LimbedRobots}). Therein, three limbs are employed to move the robot while the limb-anchored body support locomotion. These tetrahedral robots locomote with steering, forward crawling, backward crawling, in-place turning, crawling-and-turning, and rolling gaits.  In a different vein, the untethered tripodal robot proposed in \cite{tolley2014untethered} showcased jumping locomotion with the aid of combustion.

    \subsection{Quadrupedal Locomotion} \label{Subsec:Quadrupeds}

    Soft quadrupedal robots have four limbs that are designed to move in a way that mimics the movement of animals such as dogs, cats, and other four-legged creatures. This is a widely explored area within the field of soft-limbed locomotion. Soft quadrupeds are often formed by symmetrically anchoring four soft limbs to either end of a rigid body \cite{godage2012locomotion,mao2016design,li2018precharged,huang2019soft,kalin2020design,kim2021origami,tanaka2021dynamic,muralidharan2021soft,xia2021legged,gong2021untethered,atia2022reconfigurable,ji2022synthesizing,wu2022fully} or a soft body \cite{shepherd2011multigait,tolley2014resilient,faudzi2017soft,bern2019trajectory,tang2020leveraging,schiller2019toward,schiller2020gait,zhu2021quadruped,schiller2021remote,arachchige2023softsteps}. Among them, in some soft quadrupeds such as those reported in \cite{shepherd2011multigait,tolley2014resilient,mao2016locomotion,bern2019trajectory,schiller2019toward,schiller2020gait}, the limbs and the body have been molded together as a single platform. Additionally, there are soft quadruped prototypes, that have all limbs anchored to a single point or base \cite{stokes2014hybrid,mao2016locomotion,poungrat2017starfish,drotman20173d,drotman2018application,drotman2021electronics,feng2022high}. Based on the orientation of the limb arrangement, soft quadrupeds can be categorized into straight limb (i.e., keeping limbs under the body like mammals) \cite{li2021untethered} and flat or overhang limb (i.e., keeping limbs well out to the side of the body like reptiles) quadrupeds \cite{schiller2021remote}.

    A majority of flat limb quadrupeds show crawling \cite{shepherd2011multigait,godage2012locomotion,mao2016design,mao2016locomotion,poungrat2017starfish,bern2019trajectory,huang2019soft,kalin2020design,gong2021untethered,li2022scaling,wu2022fully,arachchige2023softsteps} as their fundamental mode of locomotion. They achieve it by crawling soft limbs at different phase shifts between diagonal limb pairs \cite{sun2023fully}. There are some prototypes (both straight and flat limbed robots) that show locomotion gaits such as walking \cite{tolley2014resilient,drotman20173d,faudzi2017soft,drotman2018application,li2018precharged,bern2019trajectory,drotman2021electronics,gong2021untethered,kim2021origami,muralidharan2021soft,atia2022reconfigurable,ji2022synthesizing,kaarthik2022motorized,ketchum2023automated,yang2023quadrupedal}, trotting \cite{godage2012locomotion,ansari2015dynamic,faudzi2017soft}, pacing \cite{ansari2015dynamic,xia2021legged}, back-flipping \cite{xia2021legged}, bounding \cite{xia2021legged}, undulation \cite{shepherd2011multigait}, and galloping \cite{ansari2015dynamic,tang2020leveraging, xiong2023fast}. The quadruped in \cite{wang2025body} exhibits body-induced locomotion. Its stiffness-tunable limbs \cite{huzaifa2023simplified,meng2024path} allow the body to be lifted and actuated in wave-like patterns, enabling both forward and backward motion.
    
\begin{figure*}[tb] 
		\centering
		\includegraphics[width=1\linewidth]{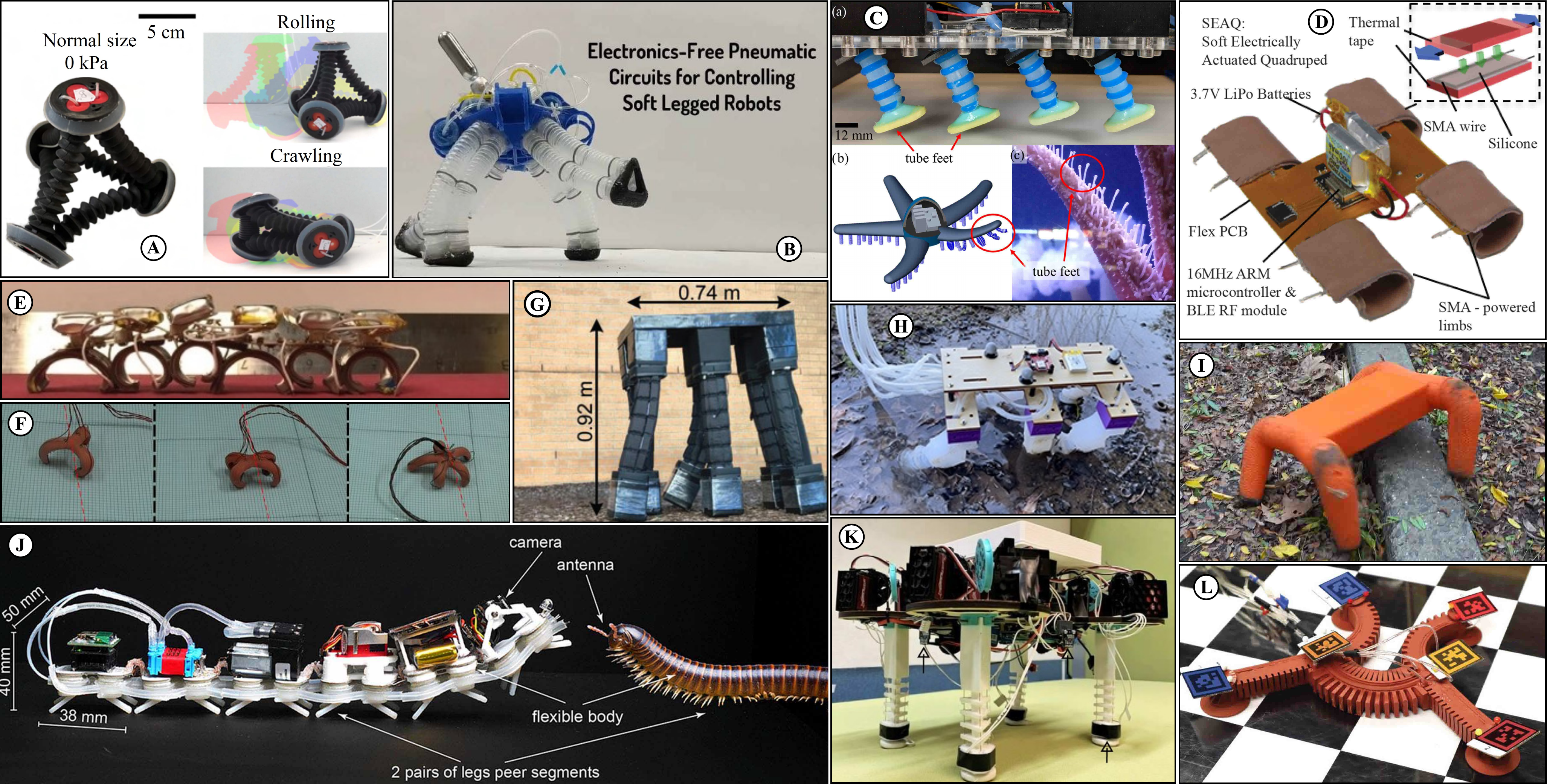}
		\caption{State-of-the-art wheelless terrestrial soft-limbed robots: (A) "Tetraflex" tetrahedral robot that replicates multimodal locomotion \cite{wharton2023tetraflex}. (B) Electronics-free quadruped in \cite{drotman2021electronics}. (C) Sea star-inspired robot that locomotes via active suction in \cite{ishida2022locomotion}. (D) "SEAQ" electrically actuated quadruped in \cite{huang2019soft}. (E) SMA actuated hexapod in \cite{goldberg2019planar}. (F) SMA actuated triped, quadruped, \& pentaped in \cite{mao2016locomotion}. (G) meter-scale hexapod in \cite{li2022scaling}. (H) "SoRX" hexapod in \cite{liu2020sorx,liu2021position}. (I) "Flexipod" motor-driven quadruped in \cite{xia2021legged}. (J) millipede-inspired multiped in \cite{shao2022untethered}. (K) Tendon-driven quadruped in \cite{muralidharan2021soft,ji2022synthesizing,ji2022omnidirectional}. (L) Gecko-inspired quadruped in \cite{schiller2019toward,schiller2020gait,schiller2021remote}. (Reproduced with permission).}
		\label{fig:Fig4_LimbedRobots} 
	\end{figure*}

    The proposed quadruped prototypes have numerous capabilities. Some can climb along parallel rods \cite{zhu2021quadruped}, walk on inclined surfaces \cite{faudzi2017soft,schiller2019toward,bern2019trajectory}, walk through height obstacles \cite{kalin2020design} or harsh environments \cite{tolley2014resilient}, carry payloads \cite{yin2019combining}, and negotiate unstructured terrains \cite{xia2021legged,drotman20173d,drotman2018application,wu2022fully}. Some quadurupeds \cite{majidi2013influence,shepherd2011multigait} use terrestrial undulation to move the robot. Readers are referred to Fig.~\ref{fig:Fig4_LimbedRobots} to see some of the state-of-the-art soft quadrupeds.

    \subsection{Pentapedal Locomotion} \label{subsec:Pentapods}

    Pentapedal robots use five legs for locomotion. Unlike traditional robots that use four or six limbs, pentapedal robots use an odd number of legs, which gives them unique advantages in terms of stability and maneuverability. There have been quite a few soft robots proposed in this topology \cite{mao2013new,mao2014gait,mao2016locomotion,jin2016soft,jin2016starfish,scott2020geometric,lee20203d}. They have soft limbs symmetrically anchored to a single point. The locomotion gaits of those pentapods proposed in \cite{mao2014gait,mao2016locomotion,jin2016starfish,scott2020geometric} are inspired by the natural locomotion of starfish. They primarily include crawling and rolling gaits \cite{scott2020geometric}. 
    Therein, they achieve continuous crawling and rolling motions by repeating the limb deformation in sequential patterns. Additionally, the five-armed actinomorphic soft robot proposed in \cite{jin2016soft} sequentially crawls its tentacle limbs to achieve locomotion. The pentapod propsoed in \cite{stokes2014hybrid} uses undulation to move the robot.

    \subsection{Hexapedal Locomotion} \label{subsec:Hexapods}
    
    Hexapod soft robots are popular since they have inherent locomotion stability by their design. In such robots, six limbs are evenly anchored to a rectangular rigid backbone \cite{suzumori1996elastic,suzumori2003applications,waynelovich2016versatile,liu2020sorx,liu2021position,li2022scaling,ishida2022locomotion}. These hexapod prototypes sequentially move limbs and achieve omni-directional walking and turning \cite{suzumori1996elastic,suzumori2003applications}, forward/backward walking \cite{suzumori1996elastic,waynelovich2016versatile,liu2020sorx,li2022scaling} gaits. Therein, all the limbs have a fundamental role in maintaining the robot’s stability. The work reported in \cite{florez2014soft} shows a hexapod that has flat limbs. It achieves crawling locomotion through limb propulsion. The hexapod proposed in \cite{calisti2012design} has limbs symmetrically anchored around a circular rigid plate and shows pushing-based crawling locomotion. The sea star-inspired hexapod appeared in \cite{ishida2022locomotion} moves the body via active suction. The meter-scaled hexapod reported in \cite{li2022scaling} achieves walking based on a locomotion pattern inspired by octopus. Some hexapod prototypes are capable of negotiating several terrains such as stone, grass, sand, mud, wood, etc. \cite{liu2020sorx,li2022scaling}. In a different vein, the authors in \cite{hase2022development,wharton2023tetraflex} proposed six-legged mobile robots that arrange their soft limbs into a closed tetrahedral shape.
    
    \subsection{Multipedal Locomotion} \label{subsec:Multipods}
    
    Herein, multi-limbed soft robots are referred to as those that have more than six limbs. The number of legs can vary depending on the specific design and application of the robot, and can range from seven legs to dozens or even hundreds of legs. It should be noted that soft multipedal locomotion is a highly under-researched area within the subfield of soft-limbed locomotion. The work reported in \cite{zou2018reconfigurable} shows a soft-bodied robot with nine reconfigurable legs. Its movements are inspired by the crawling gaits of caterpillars. Authors in \cite{shao2022untethered} proposed a soft millipede robot (Fig~\figref[J]{fig:Fig4_LimbedRobots}) driven by 24 microfluidic actuators. The work proposed in \cite{lu2018bioinspired,gu2020magnetic,lu2020battery,yang2021starfish,sun2022analysis} show magnetically actuated millirobots consisting of soft bodies and multiple tapered soft feet. These robots have shown the potential to locomote on uneven terrains and harsh environments at the cost of the complexity of controlling multiple legs. Table \ref{Table:Softlimbedtaxanomy} provides valuable insights into recent developments in soft-limbed mobile robots.

\section{Soft-Limbless (or Soft-bodied) Robot Locomotion}\label{sec:SystemModel}

Limbless soft robot locomotion refers to the movement of soft robots that do not have traditional limbs or appendages but instead rely on other means of propulsion to move. 
In the absence of limbs or appendages, the body of the soft robot essentially becomes the entire robot. Herein we discuss examples of terrestrial locomotion methods of soft-limbless robots such as peristalsis, crawling, slithering, undulation, etc. Figures~\figref[A]{fig:Fig3_NumberofPapers} and \figref[C]{fig:Fig3_NumberofPapers} show the distribution of soft-bodied robots over the years. 
    
    \subsection{Peristaltic Movements}

Peristaltic locomotion is a type of movement where the robot moves forward by squeezing or contracting its body in a wave-like motion. This type of locomotion is inspired by the way that some animals, such as worms and snakes, move. In soft robotic peristaltic locomotion, the flexible body is divided into segments, each of which can be inflated or deflated to create a wave-like motion that propels the robot forward. Such peristaltic movements result in crawling locomotion \cite{xie2024soft}. An impressive number of soft mobile robots that are based on bio-inspired peristaltic movements have been proposed to date. Among them, soft robots inspired by caterpillar \cite{trimmer2012towards,rozen2021design,sheng2020multi,goldberg2019planar,umedachi2019caterpillar,mc2020continuum,patel2023highly,henke2017soft,grossi2021metarpillar,guo2022bioinspired,wu2023caterpillar}, burrowing worm \cite{calderon2016design}, inchworm \cite{cheng2010design,cao2018untethered,duggan2019inchworm,joyee2019fully,li2019agile,wang2014locomotion,wu2018structure,tang2020design,booth2018omniskins,hu2023inchworm,zhang2021inchworm,karipoth2022bioinspired,xu2022locomotion,xu2023dynamic,dong2022bioinspired,xu2024modular}, earthworm \cite{seok2010peristaltic,seok2012meshworm,das2020soft,das2023earthworm,liu2019kirigami,manwell2018bioinspired,menciassi2004design,joey2017earthworm,nemitz2016using,onal2012origami,aydin2018design,tang2020development,du2022worm,rezaei2022earthworm}, hornworm \cite{umedachi2013highly,umedachi2016softworms}, leech \cite{huang2020multimodal}, annelid \cite{jung2007artificial,martinez2023soft}, C. elegans \cite{yuk2011shape} are featured. Some of these robots have employed passive feet \cite{sheng2020multi,sabbadini2023simple} or passive skins \cite{lee2023snakeskin} to generate anisotropic frictional forces between the body and the moving surface necessary for locomotion. Rectilinear movements of soft robotic snakes such as those reported in \cite{rafsanjani2018kirigami,rafsanjani2019propagation,lee2023snakeskin} where the robot body moves in a straight line, can also be categorized under peristaltic locomotion. One of the physical characteristics of such soft robotic snakes is that they have a shorter body length-to-diameter ratio than regular snake robots.

\begin{figure*}[tb] 
		\centering
		\includegraphics[width=1\linewidth]{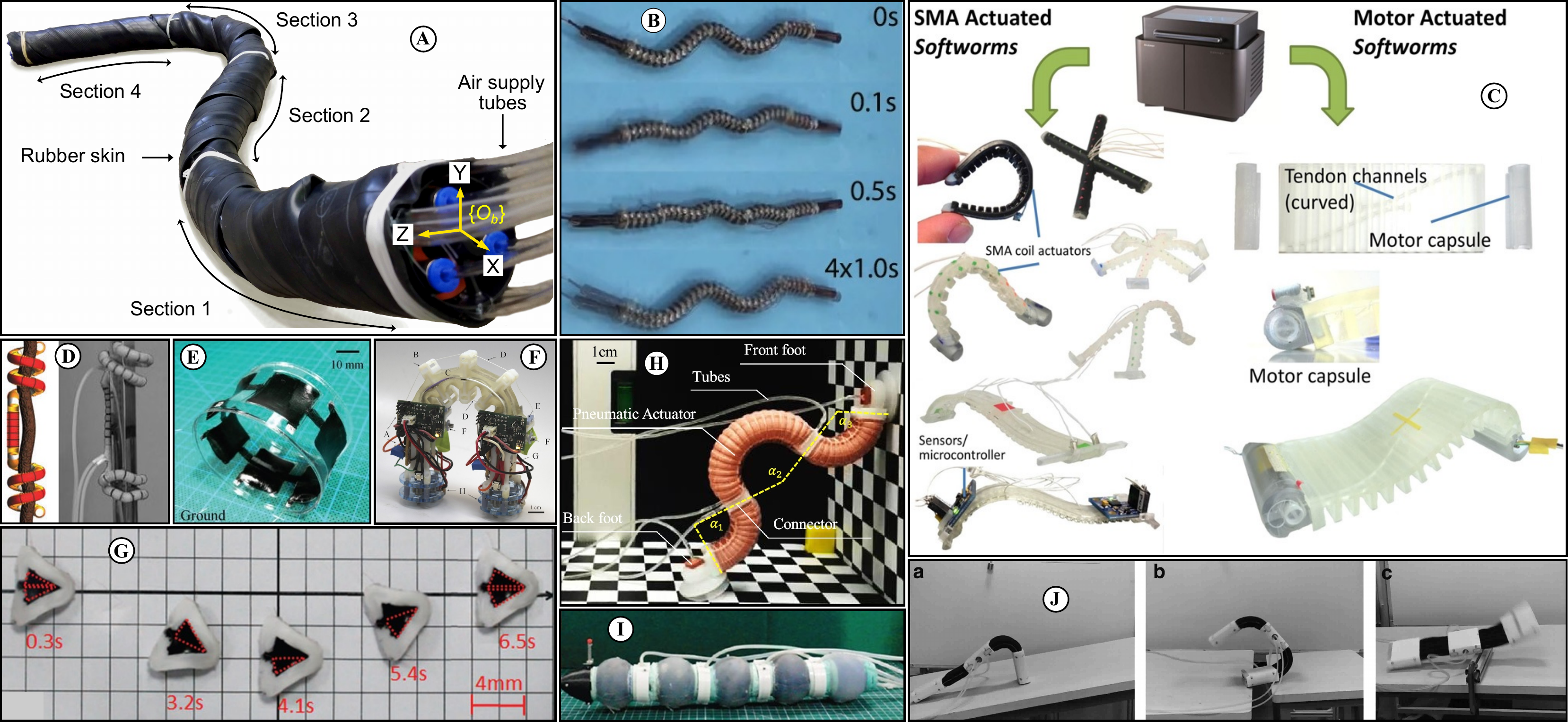}
		\caption{State-of-the-art wheelless terrestrial soft-bodied robots: (A) wheelless soft robotic snake in \cite{arachchige2023wheelless} that replicates helical rolling and sidewinding gaits. (B) Electrically actuated snake robot in \cite{zhao2021multigait} replicating its concertina gait. (C) Non-pneumatic soft worm robots in \cite{umedachi2016softworms}. (D) Winding-styled rod-climbing robot in \cite{liao2020soft}. (E) DEA-based fast rolling robot in \cite{li2021electrically}. (F) "Flippy" electromechanically actuated autonomous climbing robot in \cite{malley2017flippy}. (G) DEA-based insect scale robot in \cite{li2019agile} showing its running gait. (H) inchworm-inspired multimodal robot in \cite{zhang2021inchworm} transiting between crawling and climbing locomotion. (I) earthworm-inspired modular robot in \cite{das2023earthworm}. (J) "Fifobots" flipping robot in \cite{wang2019fifobots} replicating its folding and moving gait. (Reproduced with permission).}
		\label{fig:FIg3_LimblessSoftRobots} 
	\end{figure*}

Some other categories of soft robot peristaltic movements include wall-climbing \cite{gu2018soft,malley2017flippy,qin2019versatile,zhang2021reconfigurable,chen2023design,zhang2019modular,pratap2017wall}, amphibious climbing \cite{tang2018switchable,zhang2019multimodal,jiang2020multimodal,sakuhara2020climbing,zhang2019worm,xie2018pisrob}, worm-inspired pipe crawling \cite{liu2022worm,lin2023single,wan2022bionic,yu2022minimally,zhang2019design,chen2023design,singh2019pipe,xu2021modelling}, tube climbing \cite{verma2018soft,zhang2019worm,kanada2019reachability,hu2022soft,hu2019inchworm,li2019research,mendoza2024combined}, and mesh-worm crawling \cite{horchler2015peristaltic}. Figure~\ref{fig:FIg3_LimblessSoftRobots} shows a few state-of-the-art soft-bodied robots that rely on body peristaltic in their locomotion. 

    \subsection{Undulation} 

    Herein, soft robots mimic the undulating movements of animals such as worms, caterpillars, and snakes. They move by propagating waves along their body in a more linear fashion. To achieve it, limbless soft robots typically use some form of actuation method to change the shape or volume of their body, which generates a wave-like motion in a forward or backward direction. The undulating motion primarily depends on the anisotropic friction between the moving surface and the robot's skin \cite{li2024enhancing}. Soft robotic snakes \cite{arachchige2021soft,ta2018design,zhao2021multigait} frequently use undulation to move through planar surfaces. Moreover, soft robots inspired by C. elegans \cite{yuk2011shape} and annelid \cite{jung2007artificial} also mimic undulatory locomotion.

   \subsection{Lateral Undulation} 
    
    Inspired by biological snake movements, lateral undulation is frequently used by soft robotic snakes. Therein, the robot's body moves in a wave-like motion in a sideways (i.e., diagonal waves) or lateral direction rather than a forward or backward direction. Soft robots that use lateral undulation such as those reported in \cite{branyan2017soft,cao2017novel,branyan2020snake,qi2020novel,branyan2022curvilinear} have segmented bodies, with each segment able to bend and flex independently of the others. To move forward and sideways, the robot contracts the muscles on one side of its body while relaxing the muscles on the other side, causing the body to bend in the opposite direction. This is particularly effective for moving across smooth, flat surfaces.

  \subsection{Serpentine Movements} 
    
    Serpentine locomotion, involves the robot moving its body in a wave-like pattern from side to side, creating a series of curves along its entire length. The snake uses these curves to propel itself forward, using a combination of muscle contractions and lateral undulation. This type of locomotion is typically used by small, slender-bodied snakes, such as garter snakes, and it is particularly effective for moving quickly over relatively smooth surfaces. Similarly, soft robotic snakes proposed in \cite{qi2022bioinspired,arachchige2021soft,branyan2017soft,branyan2018soft,lopez2022muscle,haghshenas2022dynamics} have replicated serpentine gaits with promising results. It must be noted that in some soft robotic snake research \cite{branyan2017soft,branyan2018soft}, both serpentine and lateral undulations have been considered identical and tested for gait replications.

\subsection{Slithering} 

Slithering is a type of movement that involves the use of the robot's body to slide or glide along a surface, typically with the help of friction-reducing adaptations. It involves the robot's body moving in a serpentine pattern, with lateral undulations that propagate from the head to the tail. This type of locomotion is different from undulatory locomotion in that it involves a more lateral motion of the robot's body, rather than a wave-like motion. Soft robots such as those reported in \cite{branyan2017soft,cao2017novel,branyan2022curvilinear} mimic slithering locomotion. 

    \subsection{Sidewinding}

    Snakes use sidewinding locomotion when they want to minimize skin-ground contacts (such as in deserts). Inspired by that, soft robots achieve sidewinding motion, by throwing the body into a series of helices \cite{kim2024sidewinder}. For that, the robot sequentially activates different segments of the robot's body. The robot first anchors one end of its body to the ground, then contracts the muscles in the middle segments, causing them to bend and lift off the ground. The robot then moves the lifted segments forward and re-anchors them, repeating the process along the length of its body to create a continuous, wave-like motion that propels it forward. Because of lifting off the ground, sidewinding is a spatial locomotion gait. In limbless soft robots, the soft robotic snake prototypes proposed in \cite{zhao2021multigait,arachchige2023wheelless,rozaidi2023hissbot,zhou2024locomotion}, demonstrate sidewinding gaits well. 

    \subsection{Concertina Locomotion}

    Snakes use concertina movement to navigate through narrow spaces or climb obstacles. This type of locomotion involves the snake bunching up its body in alternating S-shapes, which allows it to extend its head and tail into a small opening and then pull the rest of its body through the opening. In soft robotic snakes, this is done by sequentially moving the sections of the robot's body forward and then pulling the other sections forward to meet them. The flexible robotic snake prototype proposed in \cite{zhao2021multigait}, mimics this concertina gait.

    \subsection{Rolling}

    Limbless soft robots use rolling locomotion to efficiently move their bodies sideways, cross over obstacles, etc. There are several different types of rolling soft robots, each with its unique design and locomotion mechanisms. One common type of rolling soft robot is the "rolling cylinder" which is a cylindrical robot that deforms its body to roll forward. Soft robotic snakes that show planar \cite{arachchige2021soft,arachchige2023dynamic,zhou2024locomotion} and spatial rolling \cite{arachchige2023dynamic,arachchige2023wheelless} locomotion fall into this category. Another category of bioinspired rolling includes caterpillar rolling \cite{lin2011goqbot,patel2023highly}. Further, there are soft robot prototypes such as soft-wheeler robots \cite{li2018fast,li2021electrically}, isoperimetric soft robot \cite{usevitch2020untethered}, magnetic grasping robot\cite{wang2022magnetic2} that can effectively mimic rolling locomotion. 
    
The work reported in \cite{goldberg2019planar} shows a soft-bodied robot that has a large number of wedge-shaped legs. Its movements are inspired by rolling gaits of caterpillars. Similarly, the work presented in \cite{henke2017soft} shows a caterpillar-inspired robot fabricated with six body segments and five pairs of legs. It moves with the help of body oscillations. 

\begin{table*}[hbt!] 
\setlength{\tabcolsep}{6pt} 
\renewcommand{\arraystretch}{1.4}
\caption{\textsc{Taxonomy of recent locomotion research on wheelless terrestrial soft-limbed robots}}
\label{Table:Softlimbedtaxanomy}
\begin{tabular}{lllllllll}
\toprule\toprule
Research & \begin{tabular}[c]{@{}l@{}}Locomotion \\ Method\end{tabular} & Actuation & \begin{tabular}[c]{@{}l@{}}Modeling\\ Approach\end{tabular} & \begin{tabular}[c]{@{}l@{}}Trajectory\\ Generation\end{tabular} & Control & \begin{tabular}[c]{@{}l@{}}Power \\ Autonomy\end{tabular} & \begin{tabular}[c]{@{}l@{}}Limb\\ Compliance\end{tabular} & \begin{tabular}[c]{@{}l@{}}Max. \\ Speed\end{tabular} \\ \midrule
Perera2023 \cite{perera2023teleoperation} & Tripedal  & Penumatic & CCM & Kinematic & Teleoperation & Tethered & Active & 0.65 BL/s \\
Wang2021 \cite{wang2021design} & Tripedal & Penumatic & CCM & Kinematic & Open-loop & Tethered & Active & 3.7 mm/s \\
Bern2019 \cite{bern2019trajectory} & Tripedal & Electromec. & FEM & Dynamic & Open-loop & Tethered & Active & 7.4 BL/min \\
Mao2016 \cite{mao2016locomotion} & Tripedal & SMA & -- & Custom & Open-loop & Tethered & Active & -- \\
Jin2016 \cite{jin2016soft} & Tripedal & SMA & RCM & Custom & Open-loop & Tethered & Active & 0.7 BL/s \\
Wu2022 \cite{wu2022fully} & Quadrupedal & Pneumatic & FEM & Custom & Dynamic & Tethered & Active & 0.97 BL/s \\
Atia2022 \cite{atia2022reconfigurable} & Quadrupedal & DEA & -- & Custom & Open-loop & Tethered & Active & 2.1 mm/s \\
Ji2022 \cite{ji2022synthesizing} & Quadrupedal & Electromec.& Learn-based & Learn-based & Learn-based & Tethered & Active & 0.05 m/s \\
Xia2021 \cite{xia2021legged} & Quadrupedal & Electromec.& Dynamic & Custom & Closed-loop & Untethered & Passive & 2.5 BL/s \\
Zhu2021 \cite{zhu2021quadruped} & Quadrupedal & Pneumatic& Kinematic & Custom & Open-loop & Tethered & Active & 2.52 mm/s \\
Murali2021\cite{muralidharan2021soft} & Quadrupedal & Electromec. & CCM & Kinematic & Closed-loop & Tethered & Active & 2.0 BL/s \\
Schiller2021 \cite{schiller2021remote} & Quadrupedal & Pneumatic & -- & Custom & Teleoperation & Tethered & Active & 0.8 BL/s \\
Gong2021\cite{gong2021untethered} & Quadrupedal & Hybrid & Kinematic & Path-planing & Open-loop & Untethered & Active & 37.4 mm/s \\
kim2021 \cite{kim2021origami} & Quadrupedal & Pneumatic & -- & Custom & Open-loop & Untethered & Active & 6.36 mm/s \\
Drotman2021 \cite{drotman2021electronics} & Quadrupedal & Pneumatic & -- & Custom & Open-loop & Untethered & Active & 0.13 BL/s \\
Tang2020 \cite{tang2020leveraging} & Quadrupedal & Pneumatic & FEM & Bio-inspired & Control & Tethered & Passive & 2.49 BL/s \\
Kalin2020 \cite{kalin2020design} & Quadrupedal & Electromec. & Dynamic & Dynamic & Closed-loop & Untethered & Active & 0.83 BL/s \\
Huang2019 \cite{huang2019soft} & Quadrupedal & SMA & -- & Custom & Open-loop & Untethered & Active & 0.56 BL/s \\
Li2018 \cite{li2018precharged} & Quadrupedal & Electromec. & Modeling & Custom & Open-loop & Untethered & Active & 10 mm/s \\
Drotman2017 \cite{drotman20173d} & Quadrupedal & Pneumatic & CCM & Custom & Open-loop & Tethered & Active & 0.13 BL/s \\
Poungrat2017\cite{poungrat2017starfish} & Quadrupedal & Pneumatic & -- & Custom & Open-loop & Tethered & Active & 7.0 mm/s \\
Faudzi2017 \cite{faudzi2017soft} & Quadrupedal & Pneumatic & -- & Custom & Open-loop & Tethered & Active & 0.056 BL/s \\
Mao2016 \cite{mao2016design} & Quadrupedal & Pneumatic & -- & Custom & Open-loop & Tethered & Active & -- \\
Ansari2015 \cite{ansari2015dynamic} & Quadrupedal & Pneumatic & Custom & Custom & Open-loop & Tethered & Active & 3.75 cm/s \\
Stokes2014 \cite{stokes2014hybrid} & Quadrupedal & Pneumatic & Custom & Custom & Open-loop & Untethered & Active & 6.5 m/h \\
Tolley2014 \cite{tolley2014resilient} & Quadrupedal & Pneumatic & Custom & Preprogramed & Open-loop & Untethered & Active & 18 m/h \\
Godage2012 \cite{godage2012locomotion} & Quadrupedal & Pneumatic & Dynamic & Kinematic & Open-loop & Tethered & Active & -- \\
Shepherd2011 \cite{shepherd2011multigait} & Quadrupedal & Pneumatic & -- & Custom & Open-loop & Tethered & Active & -- \\
Scott2020 \cite{scott2020geometric} & Pentapedal & Pneumatic & PDERM & T. Generation & Control & Tethered & Active & -- \\
Lee2020 \cite{lee20203d} & Pentapedal & Magnetic & -- & Custom & Open-loop & Untethered & Active & 0.25 BL/s \\
Mao2016 \cite{mao2016locomotion} & Pentapedal & SMA & -- & Custom & Open-loop & Tethered & Active & -- \\
Jin2016 \cite{jin2016soft} & Pentapedal & SMA & RCM & Custom & Open-loop & Tethered & Active & 0.7 BL/s \\
Jin2016\cite{jin2016starfish} & Pentapedal & SMA & RCM & Custom & Open-loop & Tethered & Active & 70 mm/s \\
Mao2014 \cite{mao2014gait} & Pentapedal & SMA & -- & Custom & Open-loop & Tethered & Active & 130 mm/min\\
Mao2013 \cite{mao2013new} & Pentapedal & SMA & -- & Custom & Open-loop & Tethered & Active & -- \\
Ishida2022 \cite{ishida2022locomotion} & Hexapedal & Pneumatic & Custom & Custom & Closed-loop & Tethered & Active & -- \\
Li2022 \cite{li2022scaling} & Hexapedal & Pneumatic & CCM & Custom & Closed-loop & Untethered & Active & 4.5 cm/min \\
Liu2021 \cite{liu2021position} & Hexapedal & Pneumatic &Custom & Custom & Closed-loop & Tethered & Active & 0.44 BL/s \\
Liu2020 \cite{liu2020sorx} & Hexapedal & Pneumatic &Custom & Custom & Closed-loop & Tethered & Active & 0.44 BL/s \\
Goldberg2019 \cite{goldberg2019planar} & Hexapedal & SMA & PDERM & Dynamic & Closed-loop & Untethered & Active & -- \\
Waynel2016 \cite{waynelovich2016versatile} & Hexapedal & Pneumatic & Learn-based & Learn-based & Learn-based & Untethered & Active & 40 m/h \\
Florez2014 \cite{florez2014soft} & Hexapedal & Pneumatic & Modeling & Custom & Closed-loop & Tethered & Active & -- \\
Calisti2012 \cite{calisti2012design} & Hexapedal & Electromec. & Custom & Custom & Open-loop & Tethered & Active & 20 mm/s \\
Shao2022 \cite{shao2022untethered} & Multipedal & Pneumatic & Dynamic & bioinspired & Closed-loop & Untethered & Active & 1.35 BL/s \\
Yang2021 \cite{yang2021starfish} & Multipedal & Magnetic & Modeling & Custom & Closed-loop & Untethered & Active & 5 mm/s \\
Lu2020 \cite{lu2020battery} & Multipedal & Magnetic& -- & Custom & Closed-loop & Untethered & Active & -- \\
Lu2018 \cite{lu2018bioinspired} & Multipedal & Magnetic & Custom & -- & Open-loop & Unethered & Active & 40 BL/s \\
Zou2018 \cite{zou2018reconfigurable} & Multipedal & Pneumatic & Modeling & Bioinspired & Open-loop & Tethered & Active & 2 BL/min \\
\bottomrule\bottomrule
\end{tabular}
\end{table*} 

 
    \subsection{Flipping}

    The soft robot prototypes proposed in \cite{wang2019fifobots,wang2019design,malley2017flippy} have the ability to flip the entire body and move in different directions. When activated, the flipping mechanism causes the robot's body to quickly deform and then return to its original shape, generating a flipping motion that propels the robot forward. Alternatively, the soft quadruped in \cite{xia2021legged} showed back-flipping through rapid rotation of its front and hind limbs, while maintaining a timed delay between them.

   \subsection{Winding}

   Winding is a type of locomotion used by arboreal snakes for climbing trees. Inspired by that, authors in \cite{liao2020soft} proposed a winding-styled soft rod-climbing robot (see Fig.~\figref[D]{fig:FIg3_LimblessSoftRobots}) that consists of two winding actuators and a telescopic actuator. The purpose of this movement is, anchoring the robot itself to the outer surface of a rod while the remaining body actuates and pushes the robot within the rod. A helical winding snake robot was proposed by authors in \cite{arachchige2026soft} adopting locomotion on curved surfaces.

\subsection{Jumping}

Soft-bodied robots with their compact and lightweight design can easily perform jumping \cite{chen2024bioinspired,zheng2022scalable,fernandes2022power}. The design and control of jumping mechanisms require precise tuning to manage the balance between force generation, energy efficiency, and landing stability \cite{sun2020tuning}. This form of locomotion offers unique advantages in terms of overcoming obstacles and navigating rough terrain, making it particularly appealing for applications in search and rescue, environmental monitoring, and exploration where adaptability and durability are crucial. The robot proposed in \cite{chen2021legless} is capable of fast multimodal locomotion through continuous jumping. The untethered robot reported in \cite{loepfe2015untethered} powers jumping through combustion. Table \ref{Table:Softlimbelesstaxanomy} provides insightful details on recent developments in soft-bodied mobile robots.

	\section{{Actuation Methods}\label{sec:Physics-Model}}
 This section categorizes soft mobile robots based on their actuation technologies. Therein, we discuss the pros and cons of each actuation technology and present how these technologies are related to robot shape, design, and locomotion methods.  

      \subsection{Pneumatic Actuation}

      Pneumatic actuation systems are a common method used to power and control soft mobile robots \cite{walker2020soft}. These systems use compressed air to create movement and force in the robot's flexible, deformable body. The basic components of a pneumatic actuation system for a soft mobile robot include an air compressor, a set of valves, and pneumatic actuators. The air compressor pressurizes the gas and sends it through a series of tubes to the valves, which control the flow of air to the actuators. The actuators are typically made of flexible materials, such as elastomers or silicones, and are designed to expand or contract in response to changes in air pressure.

      Pneumatic systems are lightweight and can be easily integrated into the robot's flexible body, allowing for a greater range of motion and control. Consequently, soft limbs have higher degrees of freedom (i.e. higher workspace). For example, the quadrupeds reported in \cite{suzumori1996elastic,arachchige2023softsteps,drotman20173d,wu2022fully} have 3-DoF soft limbs enabling a variety of locomotion gaits including turning. The large deformation range of the pneumatic actuators in \cite{liao2020soft,zhu2021quadruped} allowed the robot to demonstrate rod-climbing. Further, pneumatic power can effectively enable active compliance in limbs and soft bodies \cite{arachchige2023softsteps,schiller2019toward}. In a dynamic situation, actively compliant soft limbs and bodies can provide better locomotion maneuverability than passive limbs. 
      
      Pneumatic is a high-force power source that enables high-speed actuation of soft mobile robots. For example, Tang~et~\textit{al.} in \cite{tang2020leveraging} developed a soft-bodied high-speed cheetah-like robot that can gallop at a locomotion speed of 2.68 body length/s. Additionally, pneumatically powered soft mobile robots are more durable and resilient. This is proved by soft mobile robots appeared in \cite{tolleymichael2014resilient,branyan2018soft} demonstrating locomotion under extreme environmental conditions. 
      
      
      However, pneumatic actuation systems also have some drawbacks. They can be noisy and require a source of compressed air, which can limit their mobility and require additional equipment to be carried by the robot. They also require careful control and regulation of air pressure, which can be difficult to achieve in some environments. Soft mobile robots have on-board and off-board pneumatic actuation systems. On-board soft-limbed \cite{tolley2014resilient,drotman2021electronics,kim2021origami,li2022scaling} and limbless \cite{duggan2019inchworm} systems have mini air pumps fixed in the bodies making them self-contained, untethered units. But their pneumatic pressure supply is low. Hence they have a higher locomotion autonomy at the cost of low speeds \cite{drotman2021electronics}. Untethered robots are typically designed on a small scale such that low pneumatic pressure is sufficient to drive actuators. Occasionally, there might be an exception, such as the meter-scale hexapod proposed in \cite{li2022scaling}. Therein, the authors have used a vacuum pump that fits onboard and power meter-scale limbs. On the other hand, some tethered systems such as those reported in \cite{suzumori1996elastic,tang2020leveraging,arachchige2024efficient,ansari2015dynamic,liu2020sorx,shepherd2011multigait} leverage the availability of medium to high range pneumatic pressure ($\approx{1-4~bar}$) to demonstrate diverse omnidirectional locomotion patterns. Additionally, tethered limbless systems such as inchworms \cite{tang2020design,liu2022worm,duggan2019inchworm}, earthworms \cite{joey2017earthworm,liu2019kirigami,das2023earthworm}, and pipe crawlers \cite{zhang2019worm,wan2022bionic,lin2023single} achieve their peristaltic movements using low-pressure actuation.
    
    Some mobile robots are powered based on vacuum-based \cite{ishida2022locomotion,lee20203d} and prestressed \cite{xu2022pneumatic} actuation. Due to the use of negative pressure actuation and pressurized deformation, their locomotion variety and speeds are relatively low. Often their limbs require additional support to carry the body weight or payload while moving such as the one presented in \cite{ishida2022locomotion}.

 \subsection{Electromechanical Actuation}

    Electromechanical actuation is referred to the use of electric motors, tendons, and driving mechanisms such as gears for soft mobile robot actuation. There are several categories of these robots. One is tendon-driven soft mobile robots that use flexible cables to control their movement \cite{muralidharan2021soft,kalin2020design,li2018precharged,kim2023tendon,kaarthik2022motorized,ketchum2023automated}. Therein, the flexible body is segmented into multiple sections, each of which is controlled by a number of tendons that are attached to external actuators, such as motors. By pulling or releasing the tendons, the actuators can control the movement of the robot's limb or body segments, allowing it to crawl, steer \cite{manwell2018bioinspired}, undulate \cite{ta2018design}, slither \cite{malley2017flippy}, flip \cite{malley2017flippy}, or walk \cite{ji2022synthesizing} through its environment. This method of actuation is particularly suitable for soft robots as it allows for precise control of motion and can be easily integrated into the robot's overall design. However, some robots require a large number of actuators to control their movement, which can be complex and expensive to design and maintain. Additionally, the use of external actuators can limit the robot's range of motion (due to low limb dexterity such as the quadrupeds proposed in \cite{muralidharan2021soft,ji2022synthesizing}), making it difficult to perform certain tasks in confined spaces.
    
    The pipe crawling robot in \cite{yu2022minimally}, worm robot in \cite{cheng2010design} snake robots in \cite{zhao2021multigait,ta2018design}, and caterpillar-inspired soft-bodied robots in \cite{umedachi2019caterpillar,rozen2021design} are some the limbless soft robots driven by tendons. There are some robots that have soft limbs connected to a rigid body via motors at the limb-anchored joints (i.e., motorized joints). Such robots showed highly dynamic locomotion. For example, soft quadrupeds reported in \cite{kalin2020design} -- "SQuad" and in \cite{xia2021legged} -- "Flexipod" replicated multiple gaits and demonstrated locomotion in unstructured terrains. In some robots such as the pentaped reported in \cite{calisti2012design}, limbs are anchored to the motor shaft via a cranking mechanism. Another similar example; a tortoise-inspired quadruped driven by a crank shaft-rope actuation system was presented by authors in \cite{mao2016design}. One of the limitations of the above actuation methods is, limbs are actuated only at the joints (i.e. passive limbs), hence they cannot leverage active limb compliance in their dynamic locomotion. 

      \subsection{Smart Materials-based Actuation}
      Smart materials-based soft mobile robots are made of materials that have the ability to change their properties in response to external stimuli, such as temperature, light, or electric fields \cite{hao2022review}. These materials are often called "smart" because of their ability to sense and respond to their environment.
      Two popular types of smart materials and their actuation methods applied in soft mobile robots are discussed, herein.

      \subsubsection{Shape Memory Alloy-based Actuation}
      Shape memory alloy (SMA) materials have the unique ability to change shape in response to temperature changes \cite{lee2019long}. In SMA-based soft robots, the SMA material is often used in the form of wires or coils. When an electrical current is applied to the SMA wire or coil, it heats up and changes shape. As the wire cools down, it returns to its original shape, generating movement in the robot. SMA-based soft mobile robots are lightweight in nature and can produce large deformations, allowing for a wide range of motion. Leveraging such properties, the work proposed in \cite{hwang2022shape} mimicked the terrestrial crawling of seals and the work appeared in \cite{mao2013new,mao2014gait,mao2016locomotion,jin2016soft,jin2016starfish} replicated starfish locomotion.

      SMA-based robots have a slow response time due to the time required for the SMA material to heat up and change shape. Hence, they show slow locomotion as witnessed by the quadruped and the hexapod reported in \cite{huang2019soft} and \cite{goldberg2019planar}, respectively. This is further witnessed by the proposed limbless robots in \cite{mc2020continuum,menciassi2004design,onal2012origami,patel2023highly,seok2010peristaltic,wang2014locomotion} that replicated locomotion at limited speeds. It should be noted that SMA-based soft robots can be difficult to control due to the complex relationship between the input electrical signal and the resulting deformation. Owing to such characteristics, it is almost impossible to achieve dynamic locomotion with SMA-actuated soft mobile robots.


      \subsubsection{Dielectric Elastomer-based Actuation}
      Dielectric elastomers are a type of smart material that can change shape in response to an electric field \cite{guo2021review}. Typically, the elastomer material is sandwiched between two conductive electrodes. When a voltage is applied to the electrodes, the elastomer material compresses and expands, causing the robot to move.
      Atia~et~\textit{al.} in \cite{atia2022reconfigurable} proposed a legged robot based on reconfigurable dielectric elastomer actuators (DEAs). Its legs had only 1 DOF limiting locomotion speed to a low value. However, DEAs can achieve high speeds due to the fast response time of the elastomer material. For example, Li~et~\textit{al.} in \cite{li2018fast,li2021electrically} proposed rolling robots with fast speeds. Further, DEAs can have a large deformation range and can be used to make mobile robots that have a wide range of motions. For instance, the wall-climbing robot proposed in \cite{gu2018soft} shows multimodal locomotion, including climbing, crawling, and turning. DEAs are also lightweight, flexible, and can be easily fabricated into complex shapes. 

      DEAs require high voltages to generate movement, hence they have been largely used in limbless small-scale (e.g. caterpillar, worm) soft mobile robots \cite{henke2017soft,gu2018soft,guo2020all,hu2023inchworm,jung2007artificial,guo2022bioinspired} that require limited energy for deformation, hence limited mobility. Additionally, DEAs are sensitive to temperature changes and may not perform well in extreme temperatures.

       \subsection{Electrostatic Actuation}
       Here, the principle of electrostatic attraction or repulsion between oppositely charged electrodes is used to produce movement. These robots typically consist of a soft material embedded with conductive electrodes, and the application of a voltage difference between the electrodes generates an electrostatic force that deforms the soft material and produces movement. This can be seen among limbless light-weight robots such as \cite{qin2019versatile,li2019agile,cao2018untethered,chen2021legless} due to the limitations of the electrostatic force that makes the robot move. 
       
       \subsection{Magnetic Actuation}
        
        This involves the use of magnetic fields to control the movement of the robot. The magnetic particles can be either ferromagnetic or paramagnetic, and they can be dispersed in the flexible robot material randomly or in a specific pattern \cite{wang2022magnetic,lee2023magnetically}. The magnetic field is usually generated by an external source, such as an electromagnet \cite{lee20203d}, a permanent magnet \cite{joyee2019fully}, or a magnetic field generator \cite{nemitz2016using}. By changing the strength and direction of the magnetic field, the embedded magnetic particles in the robot material can be aligned or reoriented, causing the robot to deform or move in a specific direction \cite{bira2020review,eshaghi2021design,ren2024design,wang2024multimodal}. Therein, magnetic fields can be used to precisely control the movement and shape of the robot. For example, the tensegrity-based robot reported in \cite{lee20203d} showed multidirectional walking. These robots require less energy to operate, reducing the need for frequent battery replacements or recharging. However, their locomotion speed is limited due to the lack of limb or body deformation (i.e., workspace). Peristaltic robots with such characteristics have been proposed by authors in \cite{nemitz2016using,joyee2019fully,joyee20223d,ju2021reconfigurable,karipoth2022bioinspired,xu2022locomotion,ze2022soft,xu2024design}. Conversely, soft milirobots that are based on magnetic actuation \cite{lu2018bioinspired,gu2020magnetic,lu2020battery,yang2021starfish,sun2022analysis} leverage the availability of many feet to show superior locomotion performance. Additionally, some magnetic robots such as the one reported in \cite{chen2024bioinspired} show high-speed multidirectional jumping utilizing simple, lightweight actuator design. 

        \subsection{Combustion Actuation}

        The basic principle behind combustion-actuated jumping involves the controlled ignition of a fuel-air mixture within a chamber of the robot's soft body. The combustion process rapidly increases the pressure inside the chamber, causing the robot to expand and then forcefully contract, propelling it off the ground. For example, tripedal robots reported in \cite{shepherd2013using,tolley2014untethered} and soft-bodied robot reported in \cite{loepfe2015untethered} are capable of jumping powered by combustion.

       \subsection{Hybrid Actuation}
       
       There are some mobile robots that combine few actuation methods to obtain efficient movements. For instance, the work in \cite{gong2021untethered} showed a cable-driven soft quadruped made of pre-inflated pneumatic actuators. The precharged pneumatic actuator-based quadruped in \cite{li2018precharged} showed autonomous locomotion. Those robots are untethered, but their actuators show a low DoF (Mostly 2-DoF) limiting gait variations. Usevitch~et~\textit{al.} in \cite{usevitch2020untethered} proposed an untethered octahedron truss robot made of inflated fabric tubes and motorized joints. This hybrid actuation enabled them to make a human-scale mobile robot. It is capable of morphing and rolling.  The "Softworm" robots reported in \cite{umedachi2016softworms} showed SMA-actuated and tendon-driven worm robots fabricated using identical soft bodies (Fig.~\figref[D]{fig:FIg3_LimblessSoftRobots}). Moreover, the "OmniSkins" robots reported in \cite{booth2018omniskins} demonstrated rowing and inchworm locomotion using pneumatic and SMA-integrated robotic skins. An interesting hybrid actuation method has been used by the millipede reported in \cite{shao2022untethered}. It has a tendon-driven body and pneumatically actuated limbs.

\section{Modeling Approaches}

Modeling approaches applied in soft actuators (i.e., limbs and bodies) can be classified into (i) continuum mechanics models, (ii) geometrical models, (iii) discrete material models, (iv) reduced complexity models, and (v) custom analytical models \cite{armanini2023soft}.

\subsection{Continuum Mechanics Models}

Herein, actuators are defined by a configuration space that is infinite-dimensional and continuous, and they are based on the physical characteristics of the deformation. Typically, the deformation is subjected to a rigorous physical explanation of the kinetic and potential energy of the system.

    \subsubsection{Finite Element Methods}

Finite element methods (FEMs) divide the actuator into a finite number of elements (i.e., mesh in Fig.~\figref[A]{fig:Fig4_MeshandCC}) and solve equations of motion for each element based on known input parameters such as applied forces, material properties, and boundary conditions. Typically, this is implemented as a numerical simulation in soft platforms such as COMSOL, Abaqus, ANSYS, etc. In soft robot locomotion, while most of the FEM simulations are performed to predict the actuator deformation \cite{wan2022bionic,wu2022fully,zhang2019design,huang2019soft,liu2020sorx,tang2020leveraging,cao2018untethered,liu2022worm,faudzi2017soft,patel2023highly}, there are instances where FEM is employed to verify the performance of the proposed gait models. For example, Cao~et~\textit{al.} in \cite{cao2017novel} performed a FEM simulation to confirm the slithering gait of a soft robotic snake, and the results of the simulation were in close agreement with the theoretical predictions. Similarly, Bern et~\textit{al.} in \cite{bern2019trajectory} applied FEM to optimize the locomotion of a soft quadruped.

\subsubsection{Energetic Methods}

In this method, the behavior of the robot is modeled as a collection of energy storage and dissipation elements, such as springs, dampers, and other mechanical elements. The stored kinetic and potential energy is used to describe the deformation and motion of the robot. For instance, authors in \cite{patel2023highly} developed an analytic model based on the principle of minimum potential energy to predict the behavior of a bistable actuator that was used to present reconfigurable multimodal soft robots. Authors in \cite{tang2020leveraging} used an identical approach to compare the mechanical performance of a bistable spine-based actuator and its bistability-disabled counterparts. The actuator was used to assemble a high-speed crawler. 

In a different modeling perspective, energetic approaches can be used to simplify the continuous actuator models into a finite number of Lagrangian ordinary differential equations over time, while preserving the modeling's variational structure. Therein, it is assumed that the actuator is made of a set of infinitesimally thin slices with constant mass and uniform linear density. Kinetic and potential energies of a thin slice are calculated and then integrated along the length to find the total energies of the actuator \cite{godage2011dynamics}. Assume that the total kinetic energy ($\mathbf{KE}$) and the potential energy ($\mathbf{PE}$) of the actuator is known, then the complete Lagrangian can be written as \cite{godage2016dynamics}, $\mathbf{L}(\boldsymbol{q},\dot{\boldsymbol{q}})=\mathbf{KE}(\boldsymbol{q},\dot{\boldsymbol{q}})-\mathbf{PE}(\boldsymbol{q})$. Here, $\boldsymbol{q}$ defines the jointspace vector. Accordingly, the generalized equations of motion can be expressed as
\begin{align}
	\mathbf{M}\ddot{q}+\mathbf{C}\dot{q}+\mathbf{D}\dot{q}+\mathbf{K}q+\mathbf{G}+\mathbf{H}& = \mathbf{F}_{e}
\label{eq:simulatino_EoM}
\end{align}
where $\mathbf{M}$, $\mathbf{C}$, $\mathbf{D}$, $\mathbf{K}$, $\mathbf{G}$, $\mathbf{H}$, and $\mathbf{F}_{e}$ are the generalized inertia matrix, centrifugal and Coriolis force matrix, damping force matrix, elastic stiffness matrix, gravitational force matrix, hysteresis matrix, and input force vector, respectively. Using this approach, Godage~et~\textit{al.} in \cite{godage2012locomotion} expressed the dynamics of their continuum quadruped and verified its locomotion through a series of simulations. The Lagrangian approach was used by Li~\textit{et~al.} in \cite{li2018fast} and Jaryani~et~\textit{al.} in \cite{haghshenas2022dynamics} to obtain the EoM of their rolling robot and muscle-driven snake robot, respectively. Similarly, Wang~et~\textit{al.} in \cite{wang2019design,wang2019fifobots} modeled the locomotion of their flipping robots using the same approach and verified it through simulations and experiments. 

    \begin{figure}[tb] 
		\centering
		\includegraphics[width=0.8\linewidth]{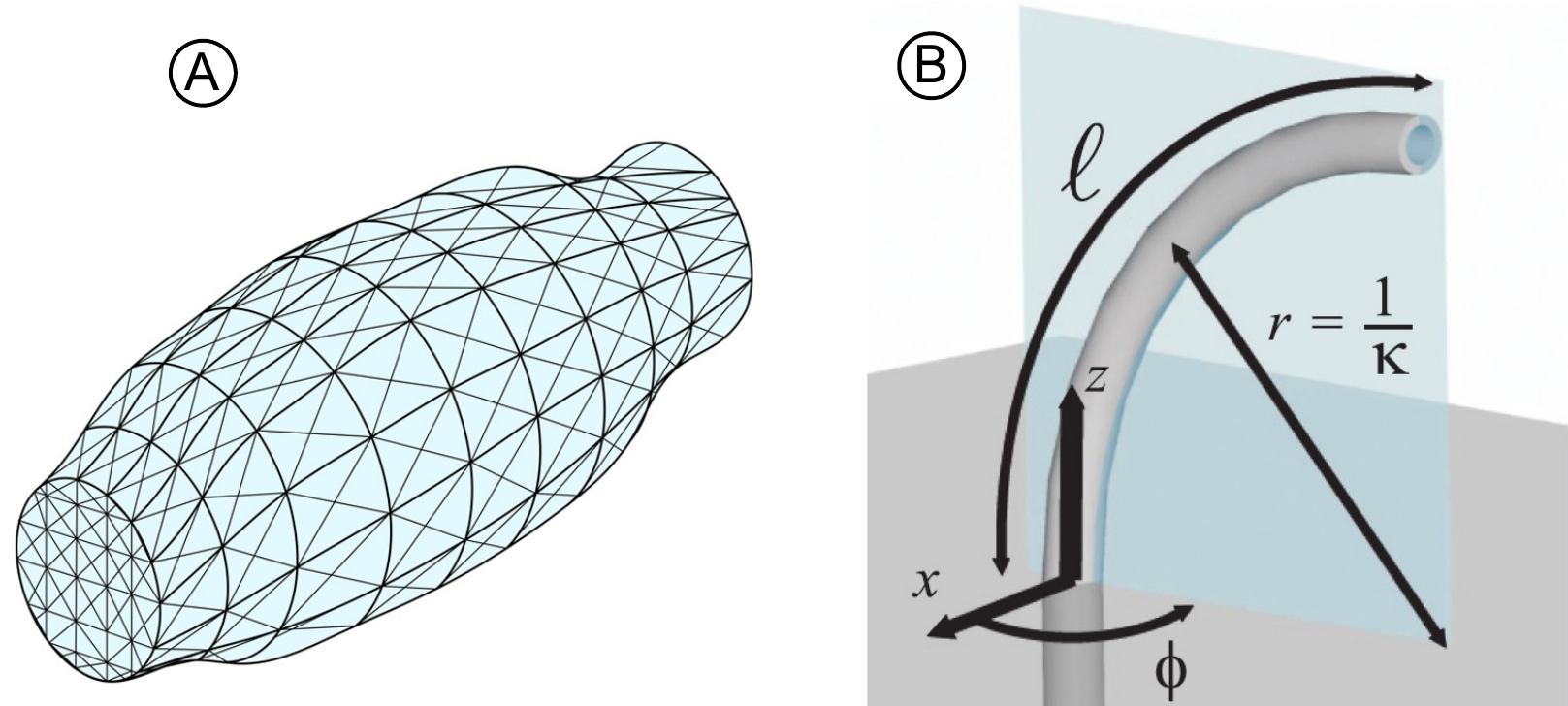}
		\caption{(A) Dividing actuator into a finite number of elements (i.e. a Mesh) in \cite{armanini2023soft}. (B) Spatial parametric representation of an actuator in \cite{webster2010design}. (Reproduced with permission).}
		\label{fig:Fig4_MeshandCC} 
	\end{figure}

\subsection{Geometrical Modeling: Constant Curvature Models}

Geometric models are based on the assumption that the deformed actuator has a resemblance to a particular geometric shape (such as the constant curvature (CC) in this case). Here, the shape is represented by a set of mathematical equations derived using curve parameters that describe the position and orientation of the actuator at any given point in time \cite{meng2021anticipatory,perera2024curve}. Typically, an actuator arc is defined by three spatial parameters; the arc length, $l$, the angle of bending plane w.r.t. the +X axis, $\phi\in [-\pi,\pi]$, and the curvature, $k$ (see Fig.~\figref[B]{fig:Fig4_MeshandCC}). Then, the transformation from arc base to any point $\xi \in [0, l]$ of the actuator is given by (note $\text{cos=c, sin=s}$)
\begin{align}
\textbf{T}(k,\phi, \xi)=\begin{bmatrix}
 \text{c}\phi \text{c}k \xi &-\text{s}\phi  &\text{c}\phi \text{s}k \xi  &\frac{1}{k}\text{c}\phi\left ( 1-\text{c}k\xi \right ) \\
 \text{s}\phi \text{c}k \xi &\text{c}\phi  &\text{s}\phi \text{s}k \xi  &\frac{1}{k}\text{s}\phi\left ( 1-\text{c}k\xi \right )  \\
 -\text{s} k \xi &0  &\text{c} k\xi  & \frac{1}{k}\text{s}k\xi \\
 0&0  &0  &1  \\
\end{bmatrix}
\label{eq:bmatrix}.
\end{align}

This CC approach has been extensively applied to model the actuators of soft tripeds \cite{arachchige2023study,perera2023teleoperation}, quadrupeds \cite{godage2012locomotion,drotman20173d,muralidharan2021soft,zhu2021quadruped,arachchige2024efficient,arachchige2023softsteps}, hexapods \cite{suzumori1996elastic,suzumori2003applications,li2022scaling}, pipe crawlers \cite{zhang2019design,yu2022minimally}, earthworm-like robots \cite{tang2020development}, and snake robots \cite{arachchige2021soft,arachchige2023dynamic,arachchige2023wheelless}. This method can be easily integrated with other modeling techniques, such as FEM or pseudo-rigid body models, to provide a more comprehensive analysis of the robot's behavior as showcased by authors in \cite{zhang2019design,yu2022minimally}.

\subsection{Discrete Modeling}

Here, the actuator is divided into discrete elements, and each element is modeled as a separate unit. Two categories of such models exist in soft mobile robot literature: pseudo-rigid body models and discrete elastic rod models.


\subsubsection{Pseudo-rigid Body Models}

Pseudo-rigid body models (PRBMs) represent the soft actuator as a series of rigid segments that are interconnected through revolute, universal, or spherical joints. They can move relative to each other while maintaining their relative angles. Thus, the actuator configuration space, $Q$ becomes
\begin{align}
Q\subset \mathbb{SE}\left(3\right) \times \mathbb{SE}\left(3\right) \times \mathbb{SE}\left(3\right)...  \times \mathbb{SE}\left(3\right).
\label{eq:se3}
\end{align}

\begin{figure}[tb] 
		\centering
		\includegraphics[width=1\linewidth]{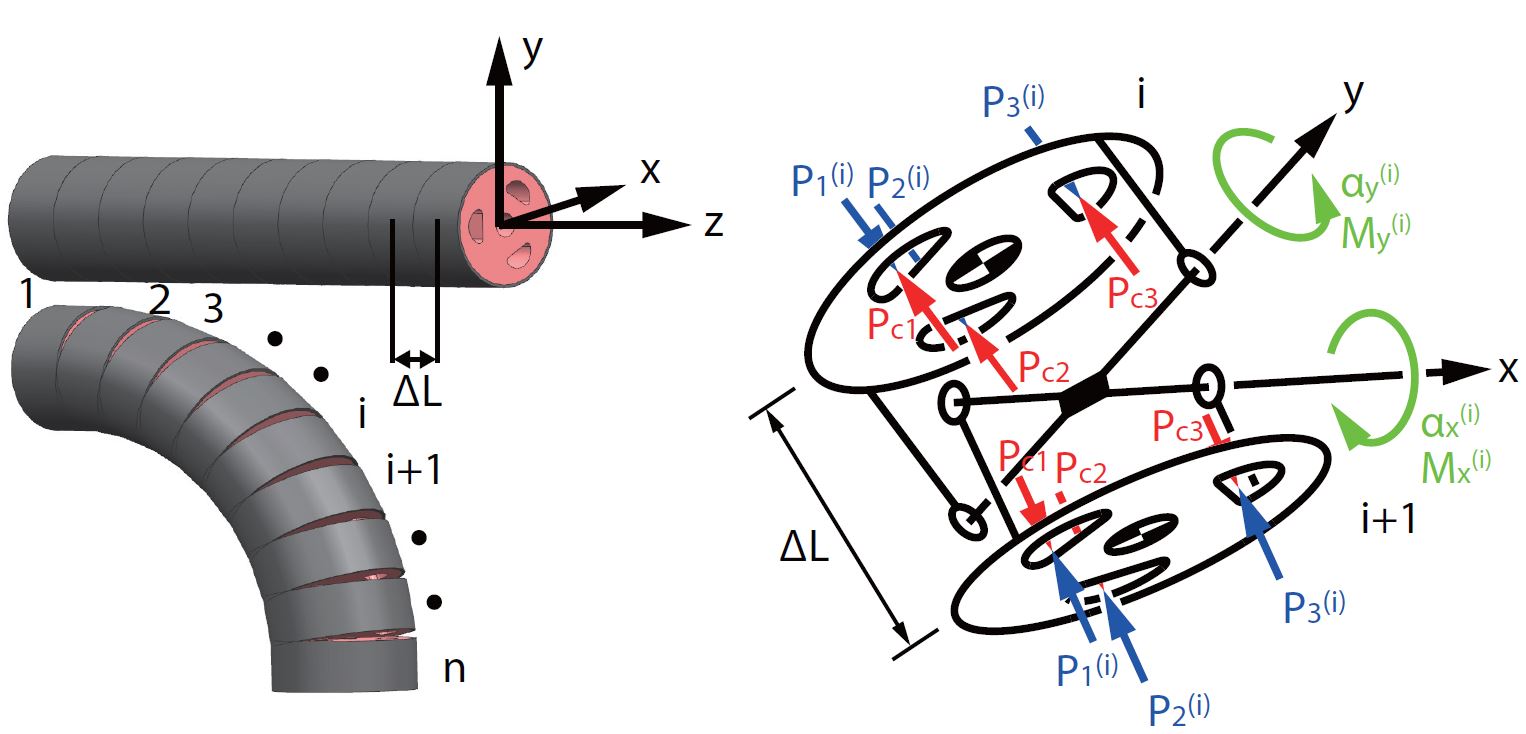}
		\caption{Left: A soft actuator of the pneumatically-powered tetrahedral robot in \cite{wang2021design} is discretized into cylindrical rigid links and cardon joints between them. Right: A simplified mechanical model of the i-\textit{th} link showing pneumatic pressure, bending torques, and joint angles. (Reproduced with permission).}
		\label{fig:Fig5_discretization} 
	\end{figure}

Authors in \cite{wang2021design} approximated a continuous soft limb of their tetrahedral robot into $n$ cylindrical rigid links. They are interconnected to each other via cardon joints (see Fig.~\ref{fig:Fig5_discretization}). Based on this simplified model, they derived the multibody dynamics of the robot and simulated it for locomotion. Li~et~\textit{al.} in \cite{li2018fast} developed a pseudo-rigid-body model to analyze the deformation and rolling mechanism of their DEA-based rolling robot. The model was simulated on ADAMS software and verified the feasibility of the proposed rolling mechanism. 

\subsubsection{Planar Discrete Elastic Rod Models}

The planar discrete elastic rod model (PDERM) represents the actuator as a series of connected line segments, or rods, that are connected by hinges. Each rod segment has a length, a bending stiffness, and a twisting stiffness, which determines how it will deform under external forces \cite{bergou2008discrete}. To simulate the motion of the robot, the PDERM uses a combination of geometric and dynamic equations. The geometric equations describe how the rods deform due to external forces, while the dynamic equations describe how the rods move under the influence of these forces. Accordingly, authors in \cite{goldberg2019planar} combined PDERM of a single SMA actuator with Lagrange’s EoM and successfully modeled the locomotion of an SMA-powered hexapod. The work on starfish-inspired robot in \cite{scott2020geometric} showcased a similar application of PDERM to obtain the dynamic equation of the full robot. 

\subsection{Reduced Complexity Models}

This refers to simplified models that capture the essential characteristics of a soft robot's behavior. Authors in \cite{joey2017earthworm,li2019agile,umedachi2019caterpillar} employed reduced-complexity dynamic models to model the crawling locomotion of their worm and caterpillar-inspired robots. Therein, they approximated the soft body and muscles into mass-spring-damper systems. Rozen-Levy~et~\textit{al.} in \cite{rozen2021design} modeled the body and the gripper dynamics of their branch-crawling worm robot based-on tendon actuated motor dynamics and spring-mass-damper dynamics of the body. Wang~et~\textit{al.} in \cite{wang2019design} developed a state-space model to simulate dynamics of its flipping robot. In a slightly different approach, the work reported in \cite{jin2016starfish,mao2014gait,jin2016soft} used thermodynamic modeling to describe the heating and cooling of SMA-based limbs. 

\subsection{Custom Analytical Models}

Several customized analytical modeling methods of actuators and locomotion gaits can be found in soft robotic locomotion research. They are unique for a particular mobile robot and its locomotion strategies.  
The magnetically actuated milli robot in \cite{yang2021starfish} obtained its dynamic model through force analysis that includes magnetic pulling force, friction force, and tapping force.

The electrically actuated snake robot reported in \cite{zhao2021multigait} was modelled based on the behavior of its individual elements such as the helical spring and the drive motor. Some work have static modeling methods such as beam theory \cite{qin2019versatile}, spring deflection equation \cite{seok2012meshworm}, Euler-Bernoulli principle \cite{huang2020multimodal,calisti2012design}, and custom geometric constraint-based model \cite{liu2021position} applied to derive the actuator deformation and then robot motion. The work reported in \cite{liu2020sorx} shows a static model based on geometric constraints for each leg



    \begin{figure*}[tb] 
		\centering
		\includegraphics[width=1\linewidth]{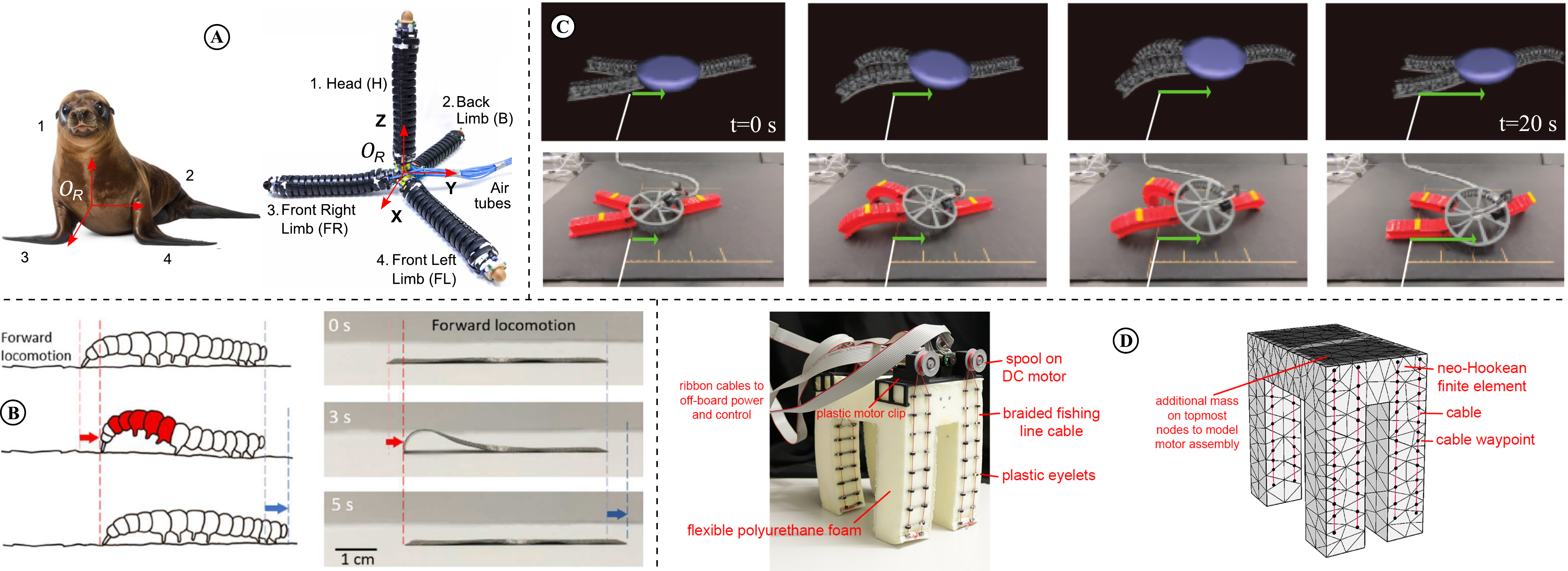}
		\caption{Representative locomotion trajectory generation methods: Robots that replicate bioinspired (A) pinniped \cite{arachchige2023study} and (B) caterpillar \cite{wu2023caterpillar} locomotion gaits. (C) Soft crawling robot that utilizes a reinforcement learning-based co-optimization framework to generate locomotion gaits \cite{schaff2023sim}. (D) A cable-driven soft robot that employes a model-based trajectory optimization method to achieve dynamic locomotion \cite{bern2019trajectory}. (Reproduced with permission).}
		\label{fig:Fig6_LocomotionTrajectoryGenerationMethods} 
	\end{figure*}

\section{Locomotion Trajectory Generation\label{sec:Experimental-Setup}}

 Trajectory generation methods are diverse, and the choice of method depends on the specific requirements of the application. They can be broadly categorized into four types: bioinspired, model-based, model-free, and custom (trial-and-error) approaches. 
       
\subsection{Bioinspired Methods}
    
    These are the approaches that are inspired by the movement patterns of animals. 
    %
    A majority of limbless robots that show peristaltic locomotion are inspired by the movement patterns of their biological counterparts. For example, worm family robots such as 
    inchworms \cite{wan2022bionic,joyee2019fully,li2019agile,karipoth2022bioinspired,cheng2010design,umedachi2013highly,zhang2021inchworm,duggan2019inchworm,hu2023inchworm,wang2014locomotion,ju2021reconfigurable,cao2018untethered}, earthworms \cite{joey2017earthworm,aydin2018design,menciassi2004design,liu2019kirigami,das2023earthworm}, burrowing worms \cite{calderon2016design}, caterpillars \cite{joyee20223d,zou2018reconfigurable,umedachi2019caterpillar,lin2011goqbot,patel2023highly,sheng2020multi,huang2020multimodal,goldberg2019planar,henke2017soft,umedachi2016softworms,rozen2021design,trimmer2012towards}, generated locomotion through a wave-like motion. To achieve this, the robot's body was made up of a series of segments or modules that contract and expand in a coordinated way, similar to the way the muscles in a real animal's body work. This motion created a wave that propagated along the length of the robot, propelling it forward. 
   Just for two specific examples for that, the caterpillar robot in \cite{joyee20223d} achieved its multimodal locomotion by sequential activation of magnetic forces in the posterior and anterior legs and the caterpillar-inspired robot in \cite{umedachi2019caterpillar}, used module contraction/bending to produce locomotion.

   Another common bioinspired method is gait analysis, which involves analyzing the way that animals move and using that information to create a mathematical model of their motion. The soft robotic snakes reported in \cite{zhao2021multigait,cao2017novel,ta2018design,arachchige2021soft,arachchige2023dynamic,arachchige2023wheelless} generated their gaits by mathematically parameterizing locomotion curves of natural snakes.
   
   There are soft-limbed robots that generate bio-inspired locomotion trajectories. The high-speed crawler reported in \cite{tang2020leveraging} actuates its bistable hybrid body and imitates the galloping gaits of high-speed cheetahs. The  millipede robot in \cite{shao2022untethered} operates its legs in 4 phases and mimics the locomotion of natural millipedes. The sea star-inspired robot in \cite{ishida2022locomotion} utilizes active suction to locomote similar to adhesive appendages of sea stars. Note that, even though some robots mimic the movement patterns of their biological counterpart, their trajectory generation methods are not bioinspired \cite{arachchige2023study}. Refer to Table I and II for more details. 

\subsection{Model-based Methods}
This involves using mathematical models to plan and generate motion trajectories for the robot. These models typically describe the robot's kinematics, dynamics, etc., of its motion, as well as the constraints and requirements of the task at hand.

     \subsubsection{Kinematic Modeling-based Methods}

kinematics can provide a relatively accurate description of the actuator deformation (see Sec. V-B). Relying on that, authors in \cite{godage2012locomotion,arachchige2023softsteps,arachchige2023study,arachchige2024efficient,perera2023teleoperation} employed curve parametric inverse kinematics of limbs to generate  locomotion trajectories of quadrupeds and tetrahedral robots. Therein, first, they defined the fundamental trajectory shape of limbs in the taskspace and then transformed the shape into joint variables and actuated the limbs according to predetermined gait patterns. In \cite{arachchige2023softsteps}, authors experimentally validated the CC-based kinematic model with higher accuracy. A similar approach has been applied to generate undulation, sidewinding, and rolling gaits of the snake robots reported in \cite{arachchige2021soft,arachchige2023dynamic,arachchige2023wheelless}. Therein, the trajectories have been generated based on an optimization-based inverse kinematic method applied to multisection snake robots. Moreover, the tendon-driven quadruped in \cite{muralidharan2021soft} used curve parametric kinematics to calculate the required length change of three control wires in each limb to obtain the desired limb deformation in locomotion.  

In some robots, the movements were assisted by custom-derived kinematics of the actuators. They were not necessarily based on curve parametric kinematics. For example, the limb kinematics of the soft crawling and walking quadrupeds in \cite{gong2021untethered} helped develop a locomotion path planning method.



\subsubsection{Optimization-based Methods} 
Optimization methods can be used to find optimal gaits that minimize a cost function, such as energy consumption or travel time. These methods typically rely on robot dynamic models. Authors in \cite{bern2019trajectory} proposed a model-based trajectory optimization method for a cable-driven soft quadruped. Therein, first, they modeled the robot using FEM. Then, forward dynamic simulations were used to predict the robot's movements and later improve it to automatically generate optimal locomotion trajectories. Arachchige~et~\textit{al.} in \cite{arachchige2024efficient} generated trotting gaits of a quadruped based on limb kinematics. Then, they simulated and optimized the gait parameters on a Physics engine-based dynamic environment. Based on optimized gait parameters, the gaits were then experimentally validated on the quadruped prototype.

     \subsubsection{Path Planning-based Methods}
     Path planning involves generating a sequence of desired poses or configurations that the robot must follow to reach its intended destination. This can take into account the compliant nature of the robot's body and its interactions with the environment. It must first define a set of goals and constraints that the robot must satisfy. The path planning algorithm then generates a set of feasible paths that satisfy the goals and constraints. Gecko-inspired quadruped reported in \cite{schiller2020gait} utilized this approach to negotiate multiple goal positions on a flat surface. Its constraints included target position, joint variables, walking speed, and turning speed. After mapping velocity space variables (i.e. speeds) to taskspace, the robot was able to recursively generate optimal references to reach a given target position. This crawling quadruped in \cite{gong2021untethered} arrived at a particular location by completing a predetermined number of cycles, which were calculated based on the known connection between the rotational angle of its motor and the distance it moves during a single motion cycle. The robot described in \cite{mahendran2023multi} employs a data-driven optimization approach to develop locomotion gaits, learning from environmental interactions. Its control architecture combines high-level and low-level controllers with real-time overhead webcam localization, allowing it to navigate obstacle-laden environments and adapt paths dynamically.
     

\subsection{Model-free (or Learning-based) Methods }
This involves generating motion trajectories without relying on explicit mathematical models of the robot or its environment. Instead, it relies on data-driven methods such as machine learning, where the robot learns to optimize its motion through trial-and-error based on experience. Despite much work on machine learning-based deformation modeling of soft actuators \cite{kim2021review,wang2021survey}, there have been few involved with soft mobile robots. Ji~et~\textit{al.} in \cite{ji2022synthesizing} utilized deep reinforcement learning (RL) to synthesize the optimal walking gaits of their quadruped. Authors in \cite{wu2018structure} proposed an inchworm-inspired differential drive robot and utilized radial basis function (RBF) neural networks to train the robot's turning capabilities on a variety of surfaces. It resulted in a mathematical model that describes the motion. Their RBF model was experimentally verified with high accuracy. 

A slightly different learning approach was used to obtain locomotion of the hexapod reported in \cite{waynelovich2016versatile}. The authors used a simulation software package that offers the user separate learning and execution routines of the robot. While in the learning mode, the operator can control the individual leg tubes using a graphical user interface, which enables the customization and optimization of a specific gait. After achieving a desirable gait, the process was stored in memory and used in the robot.

\begin{table*}[hbt!] 
\centering
\setlength{\tabcolsep}{9pt} 
\renewcommand{\arraystretch}{1.4}
\caption{\textsc{Taxonomy of recent locomotion research on wheelless terrestrial soft-bodied robots}}
\label{Table:Softlimbelesstaxanomy}
\begin{tabular}{lllllllll}
\toprule\toprule
Research & \begin{tabular}[c]{@{}l@{}}Locomotion \\ Method\end{tabular} & Actuation & \begin{tabular}[c]{@{}l@{}}Modeling\\ Approach\end{tabular} & Trajectory& Control & \begin{tabular}[c]{@{}l@{}}Power \\ Autonomy\end{tabular}  & Speed \\ \midrule
Arachch2023 \cite{arachchige2023wheelless} & Sidewinding & Pneumatic & CCM & Snake & Open-loop & Tethered & 2.1 cm/s \\
Zhao2021 \cite{zhao2021multigait} & Concertina & Electromec. & Dynamic & Snake & Closed-loop & Tethered & -- \\
Ta2018 \cite{ta2018design} & Undulation & Electormec. & Kinematic & Snake & Open-loop & Tethered & -- \\
Cao2017 \cite{cao2017novel} & Slithering & Pneumatic & Dynamic & Snake & Open-loop & Tethered & --  \\
Branyan2020 \cite{branyan2020snake} & Undulation & Pneumatic & Custom & Snake & Open-loop & Tethered & -- \\
Liao2020 \cite{liao2020soft} & winding & Pneumatic & Custom & Snake & Open-loop & Tethered & --  \\
Huang2020 \cite{huang2020multimodal} & Peristaltic & Actuation & Modeling & Leech & Open-loop & Tethered & 2.5 mm/s \\
Usevitch2020 \cite{usevitch2020untethered} & Rolling & Hybrid & Kinematics & Custom & Closed-loop & Untethered & 216 m/h \\
Wng2019 \cite{wang2019design} & Flipping & Pneumatic & Dynamic & Custom & Closed-loop & Tethered & -- \\
Li2018 \cite{li2018fast} & Rolling & DEA & PRBM & Custom & Open-loop & Tethered & 41.2 mm/(s.g) \\
patel2023\cite{patel2023highly} & Rolling & SMA & FEM & Bioinspired & Open-loop & Tethered & 1.5 BL/s \\
Li2021\cite{li2021electrically} & Rolling & DEA & Custom & Custom & Open-loop & Untethered & 1.19 BL/s \\
Calderon2016 \cite{calderon2016design} & Peristaltic & Pneumatic & Custom & Worm & Open-loop & Tethered & 2.35 mm/s \\
Xu2022 \cite{xu2022locomotion} & Peristaltic & Magnetic & Dynamic & Worm & Open-loop & Tethered & -- \\
Seok2012 \cite{seok2012meshworm} & Peristaltic & Electromec. & -- & Worm & Open-loop & Tethered & -- \\
Horchler2015 \cite{horchler2015peristaltic} & Peristaltic & Actuation & -- & Worm & Open-loop & Tethered & 2.15 mm/s \\
Seok2010 \cite{seok2010peristaltic} & Peristaltic & DEA & Custom & Worm & Open-loop & Tethered & 2.75 mm/s \\
Nemitz2016 \cite{nemitz2016using} & Peristaltic & DEA & -- & Worm & -- & Tethered & 2.11 mm/s \\
Yuk2011 \cite{yuk2011shape} & Peristaltic & SMA & Custom & C. elegan & Open-loop & Untethered & 0.85 mm/s \\
Jung2007\cite{jung2007artificial} & Undulation & Pneumatic & Custom & Annieled & Open-loop & Tethered & 1 mm/s \\
Joyee2022 \cite{joyee20223d} & Peristaltic & Magnetic & Modeling & Caterpiller & Closed-loop & Untethered & 1.23 BL/s \\
Rozen2021 \cite{rozen2021design} & Peristaltic & Electromec. & Custom & Caterpiller & Open-loop & Tethered & --  \\
Huang2020\cite{huang2020multimodal} & Peristaltic & Pneumatic & CCM & Caterpiller & Open-loop & Tethered & 1.25 cm/s \\
Umedachi2019 \cite{umedachi2019caterpillar} & Peristaltic & Electromec. & Dynamic & Caterpiller & Closed-loop & Tethered & 1.5 cm/s \\
Zou2018 \cite{zou2018reconfigurable} & Peristaltic & DEA & CCM & Caterpiller & Open-loop & Tethered & 1.1 mm/s \\
Umedachi2016 \cite{umedachi2016softworms} & Peristaltic & Electr \& SMA. & -- & Caterpiller & Open-loop & Untethered & --\\
Das2023 \cite{das2023earthworm} & Peristaltic & Pneumatic & Custom & Earthworm & Open-loop & Tethered & 1.2 cm/s \\
Liu2019 \cite{liu2019kirigami} & Peristaltic & Magnetic & Custom & Earthworm & Open-loop & Untethered & 0.86 cm/s \\
Aydin2018 \cite{aydin2018design} & Peristaltic & Pneumatic & -- & Earthworm & Open-loop & Tethered & -- \\
Joey2017 \cite{joey2017earthworm} & Peristaltic & Pneumatic & Dynamic & Earthworm & Closed-loop & Tethered & 2.4 cm/s \\
Karipoth2022\cite{karipoth2022bioinspired} & Peristaltic & SMA & Custom & Inchworm & Open-loop & Tethered & -- \\
Zhang2021 \cite{zhang2021inchworm} & Peristaltic & Magnetic & CCM & Inchworm & Closed-loop & Untethered & 2.5 mm/s \\
Duggan2019 \cite{duggan2019inchworm} & Peristaltic & Pneumatic & PRBMs & Inchworm & Open-loop & Tethered & 1.25 cm/s \\
Hu2023 \cite{hu2023inchworm} & Peristaltic & DEA & CCM & Inchworm & Open-loop & Tethered & 1.1 mm/s \\
Ju2021 \cite{ju2021reconfigurable} & Peristaltic & Pnuematic & CCM & Inchworm & Open-loop & Tethered & -- \\
Cao2018 \cite{cao2018untethered} & Peristaltic & Magnetic & PRBM & Inchworm & Open-loop & Untethered & 2.5 mm/s \\
Zhang2019 \cite{zhang2019design} & Peristaltic & Pneumatic & Custom & Pipe-crawller & Open-loop & Tethered & 11.5 mm/s \\
Yu2022 \cite{yu2022minimally} & Peristaltic & Pneumatic & Custom & Pipe-crawller & Open-loop & Tethered & 3.35 mm/s \\
Lin2023 \cite{lin2023single} & Peristaltic & Pneumatic & Custom & Pipe-crawller & Open-loop & Tethered & 29.5 mm/s \\
Verma2018 \cite{verma2018soft} & Peristaltic & Pneumatic & Custom & tube-climber & Open-loop & Tethered & 25.2 mm/s \\
Zhang2019 \cite{zhang2019worm} & Peristaltic & Pneumatic & Custom & tube-climber & Open-loop & Tethered & -- \\
Zhang2019 \cite{zhang2019worm} & Peristaltic & Pneumatic & Custom & tube-climber & Open-loop & Tethered & -- \\
Zhang2019 \cite{zhang2019worm} & Peristaltic & Pneumatic & Custom & tube-climber & Open-loop & Tethered & -- \\
Zhang2019 \cite{zhang2019worm} & Peristaltic & Pneumatic & Custom & tube-climber & Open-loop & Tethered & -- \\
Zhang2019 \cite{zhang2019worm} & Peristaltic & Pneumatic & Custom & tube-climber & Open-loop & Tethered & -- \\
Zhang2019 \cite{zhang2019worm} & Peristaltic & Pneumatic & Custom & tube-climber & Open-loop & Tethered & -- \\
Zhang2019 \cite{zhang2019worm} & Peristaltic & Pneumatic & Custom & tube-climber & Open-loop & Tethered & -- \\
Gu2018 \cite{gu2018soft} & Peristaltic & Pneumatic & Custom & wall-climber & Open-loop & Tethered & 4.5 mm/s \\ \bottomrule \bottomrule

\end{tabular}
\end{table*}

\subsection{Custom (or trial-and-error) Trajectory Generation Methods}

    Custom methods have been adopted to obtain the best possible trajectory generation method that fits into a specific robot design. A few recent works are discussed here. The motor-actuated, passive limbs enabled quadruped in \cite{xia2021legged} generated gaits by a template, where the motor position is sectioned into multiple phases at a constant velocity. The pipe crawling robot reported in \cite{yu2022minimally} generated its trajectories through a trial-and-error experimental procedure. They actuated the body in a left and right bending pattern and the locomotion was aided by uneven forward friction caused by dissimilar mechanical characteristics in the direction of its length. 
    The Multigait tetrapod reported in \cite{shepherd2011multigait} generated crawling and undulation by pressurizing limbs at different sequences. The rod climbing quadruped in \cite{zhu2021quadruped}, the tube climbing robot in \cite{verma2018soft}, the hexapod in \cite{florez2014soft}, and the high-speed rolling robot in \cite{li2021electrically}, accomplished their movements through pure experimentation. In \cite{mc2020continuum}, it was demonstrated that a basic, straight-line increase in frequency can produce a powerful wave of energy that propels the robotic caterpillar forward. The "Flippy" robot in \cite{malley2017flippy} (Fig.~\figref[F]{fig:FIg3_LimblessSoftRobots}) generated its flipping and climbing gaits utilizing a custom-designed state machine that has states: flipping, attaching, and detaching of each side. "SoRX" hexapod in \cite{liu2020sorx} employed an alternating tripod gait that guaranteed the center of mass stays within the area of support formed by the three legs that are in contact with the ground. 
    
    In some work, researchers relied on extensive modeling of the actuator (i.e., limb or body) to achieve their locomotion. Multimodal crawler \cite{patel2023highly}, bionic omnidirectional wall and tube climbing robots; "BOWR", "BOTR" \cite{huang2020multimodal}, and walking quadruped \cite{atia2022reconfigurable} are some of them. The idea behind it was, first identify the deformation characteristics of the actuator, then utilize them to show some meaningful work such as locomotive robots. More work on various custom trajectory generation methods can be found in Tables I and II.

 \section{Control Methods\label{sec:ControlMethods}}

The control methods are proposed to regulate the movement of the robot. They involve controlling the actuation media or the system (i.e., pneumatic, motor, temperature, electricity, etc.) that makes the limbs and the body move. We briefly discuss major control methods exist in wheelless terrestrial soft mobile robots. 
    
    \begin{figure*}[tb] 
		\centering
		\includegraphics[width=1\linewidth]{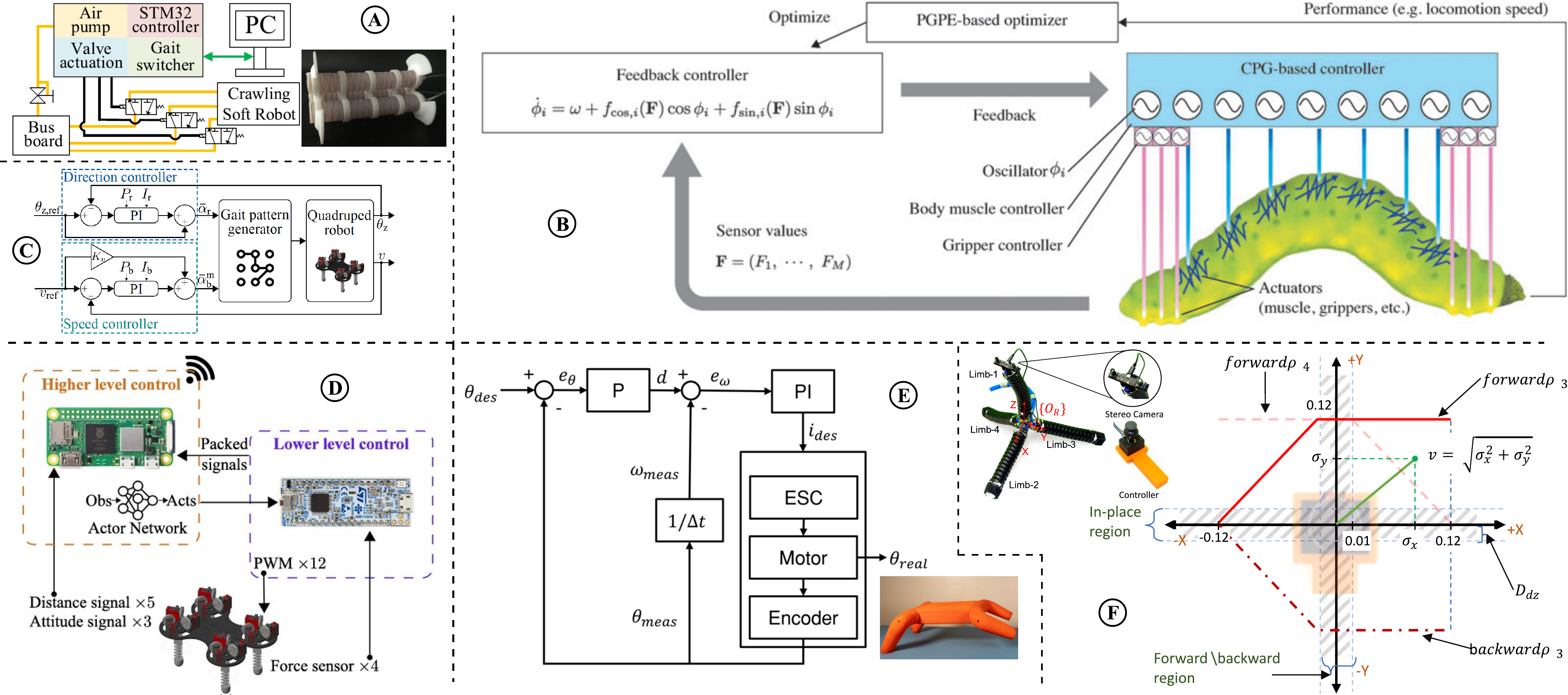}
		\caption{Representative control methods: (A) Open loop control method employed by the soft crawler in \cite{zhang2019design}. (B) Closed-loop control strategies of  in  \cite{ishige2019exploring}, (C) Closed-loop direction and speed controllers proposed by the soft quadruped in \cite{ji2022omnidirectional}. (D) Control scheme used to validate learning-based locomotion trajectories of the quadruped in \cite{ji2022synthesizing}. (E) Classical joint-space position control system used by "Flexipod" quadruped in \cite{xia2021legged}. (F) Teleoperating scheme applied to control the tetrahedral robot in \cite{perera2023teleoperation}. (Reproduced with permission)}
		\label{fig:Fig9_RepresentativeControlMethods} 
	\end{figure*}

     \subsection{Open Loop Control}

This is a type of control system where the robot's movements are pre-programmed and not dependent on feedback from sensors or the environment. In other words, the control system generates a sequence of pre-defined commands that the robot executes without taking into account any feedback or changes in the environment. One of the advantages of this method is its simplicity. Because of that, it is very popular in soft mobile robot research (see Tables I and II). Among them, the majority operated on open-loop sequential control systems \cite{mao2013new,jin2016starfish,mao2014gait,mao2016locomotion,poungrat2017starfish,li2021electrically,drotman20173d}. The quadrupeds in \cite{drotman2021electronics,shepherd2011multigait} used manual control of pneumatic valves to regulate the limb pressure. The control system of the quadruped that climbed parallel rods in \cite{zhu2021quadruped} was merely regulating air pressure. The resilient quadruped in \cite{tolleymichael2014resilient} demonstrated pre-programmed locomotion. The snake robot in \cite{zhao2021multigait} and the rolling robot in \cite{li2018fast} leveraged open-loop control to  exhibit multigait and fast locomotion, respectively. It was noted that the robots with peristaltic movements (Eg. \cite{wan2022bionic,yu2022minimally,verma2018soft,qin2019versatile,li2019agile,umedachi2019caterpillar,cheng2010design,zhang2019design}) tend to rely more on open-loop control methods. That is because they do not involve large body deformations in their locomotion. On the negative side, because the robot's movements are pre-programmed and do not depend on sensor feedback, they may be less adaptable to changes in the environment. 
     
    \subsection{Closed-loop Control}

    In closed-loop control (or feedback control), the robot's movements are regulated based on feedback from sensors and the environment. That is, the control system continuously monitors the robot's movements and environment and adjusts the robot's movements in real-time to achieve a desired outcome. This concept can be understood by observing the pressure control system (see Fig.~\ref{fig:Fig9_RepresentativeControlMethods}) employed by authors in \cite{zhang2021inchworm} who proposed a multimodal worm robot (Fig.~\figref[H]{fig:FIg3_LimblessSoftRobots}). In Fig.~\ref{fig:Fig9_RepresentativeControlMethods}, the sensor collects pressure data in various channels of the robot body in real-time and sends it back to compare with the input pressure (or desired pressure). Next, the controller receives the error signal in order to generate control signals, which in turn regulate the air pressure in the robot body. We discuss several sub-categories of closed-loop control systems hereunder.

\subsubsection{Kinematic-based Closed-loop Control Systems} 
    One approach to controlling soft robot locomotion is known as model-based kinematic control. This approach involves creating a mathematical model of the robot's kinematics and using it to design control algorithms that can accurately control the robot's movements. 

    The large-scale hexapod appeared in \cite{li2022scaling} implemented a piecewise constant curvature-based kinematic model to control the limb curvature (i.e., bending). Further, they adopted a PID regulator to address issues related to the saturation of the actuators and unmodeled interactions with the environment. The tendon-driven limbed enabled quadruped in \cite{muralidharan2021soft} implemented a kinematic controller powered by linear–quadratic regulator (LQR) to regulate the rotational position of motors that drive tendons. The isoperimetric robot in \cite{usevitch2020untethered} implemented a kinematic-based proportional-integral-derivative (PID) controller to drive the robot's rollers to the desired position. It must be noted that kinematic-based control is limited by the mathematical models used to describe the robot's motion. These models may not be able to accurately capture the full range of robot movements, particularly in complex or dynamic environments.

\subsubsection{Dynamic-based Closed-loop Control Systems (DbClCS)} 
    Here, first, mathematical models that describe the robot's movement dynamics are obtained. They are then applied to design control algorithms that generate commands for the robot's actuators, to achieve desired motions. Similar to kinematic control, these algorithms are employed in a feedback loop that adjusts the control inputs in real-time. For example, the tortoise-inspired robot in \cite{wu2022fully} applied an empirical dynamic model to optimize motion speeds. The earthworm-inspired robot in \cite{joey2017earthworm} implemented a friction-based feedback control strategy that enables active friction control and supports locomotion. Therein, the authors demonstrated how the robot attains movement by employing time-varying friction controlled through feedforward mechanisms. "Flexipod" quadruped in \cite{xia2021legged} and "SQuad" quadruped in \cite{kalin2020design} used a classical joint-space position control system (a full state feedback controller) based on robot dynamics to control the motor shaft position that enabled the limb movements. Note that, dynamic-based control systems often require significant computational resources to solve the equations of motion and control the robot in real-time.

\subsubsection{Custom Closed-loop Control Systems} 

    This category of closed-loop control involves regulating the robot's movement without the kinematics or dynamics of the robot. This method has been widely applied on a variety of occasions. For example, the quadruped that has movements enabled by active suction in \cite{ishida2022locomotion}, used a closed-loop control system to maintain the quality of suction (i.e., negative pressure). Therein, the objective was to control the vacuum quality inside the actuators and  maintain uniform movements. Authors in \cite{liu2021position} successfully applied a closed-loop trajectory tracking system to move their hexapod robot in a desired path on a flat terrain. They used an onboard motion-capturing system to sense the robot's current location and actuate the limbs such that the robot maintains its desired path. The hexapod appeared in \cite{florez2014soft} used a proportional-integral (PI) controller combined with an inertial measurement unit (IMU) to regulate its movements. The starfish-inspired robot in \cite{scott2020geometric} applied pressure and feedback control to its locomotion utilizing pressure sensors and visual tracking of the limb-tip. The gecko-inspired quadruped in \cite{schiller2020gait} demonstrated a closed-loop position control scheme in the cartesian space. Flipping robots; "FifoBots" reported in \cite{wang2019fifobots} tracked the desired flipping path with the aid of a closed-loop feedback control system that corrects the robots’ position and orientation. The robot in \cite{karipoth2022bioinspired} demonstrated cyclic closed-loop-controlled inch-worm and earth-worm locomotion. The robot utilized intrinsic strain sensors that gave instantaneous feedback to manipulate the locomotion. Manwell~et~\textit{al.} in \cite{manwell2018bioinspired} used a closed-loop controller (PID) to regulate the body contraction of their worm robot. These closed-loop control strategies showed the potential for task-oriented path planning as well. 

    \subsection{Teleoperation (or Remote control)}

    Teleoperation allows an operator to control the movement of a soft robot from a remote location, using a computer, joystick, or other input device. This typically requires a communication link (wires, Wi-Fi, bluetooth, ethernet, etc.) between the robot and the operator. It can be tethered or tetherless. 

\subsubsection{Tethered Teleoperation} 

In this method, a physical connection (tether) between the robot and a remote operator exists. The operator controls the robot's movements by sending signals through the tether, which is typically a wire that is connected to the robot. The tether provides power and communication between the robot and the operator, allowing the operator to see through the robot's sensors and control its movements in real-time. Perera~et~\textit{al.} in \cite{perera2023teleoperation} showcased the teleoperation of a tetrahedral robot with real-time stability and trajectory control. Authors in \cite{schiller2021remote} developed an intuitive user interface to remotely operate their Gecko-inspired quadruped. This approach can be useful for robots that may not be able to navigate complex environments on their own, as the tether can provide additional support and stability. However, the tether can also limit the range of movement and mobility of the robot, making it less suitable for certain applications.

\subsubsection{Untethered Teleoperation} 

    Here we control the robot without the use of a physical connection (tether). Magnetically actuated robots such as those appeared in \cite{lu2018bioinspired,lu2020battery} are categorized under tetherless teleoperating robots because the magnetic field that causes to move the robot can be remotely controlled. The millipede in \cite{shao2022untethered} was designed with tetherless remote-controlling capabilities. Its motion is assisted by real-time images sent back wirelessly. The hybrid robotic system in \cite{stokes2014hybrid}, "Flexipod" quadruped in \cite{xia2021legged}, are some examples of tetherless teleoperating systems. One of the main advantages of these methods (tethered or tetherless) is that it allows for human supervision and decision-making. Since the operator is in control of the robot's movements, they can make decisions in real-time based on the information provided by the robot's sensors. This can be particularly useful in situations where the robot encounters unexpected obstacles or challenges.

    \subsection{Learning-based Control}
    
    The control systems can also involve machine learning algorithms that enable the robot to learn from its environment and adapt its movements accordingly. For example, a soft robot might use reinforcement learning to learn how to move through a complex environment, such as a cluttered room or a maze. In \cite{ji2022synthesizing} and \cite{schaff2023sim}, learned RL-based controllers are applied to optimize the walking and crawling gaits of quadrupeds, respectively. Therein, first, the controller is trained in a simulation environment. Then, the learned gait control policy was successfully implemented and tested on the quadruped prototypes. On the contrary, the work in \cite{ketchum2023automated} utilizes a simulation-free tree-search-based method to automatically generate walking gatis.

  \section{Research Challenges\label{sec:ResearchChallenges}}

Despite the substantial amount of research and advancements reviewed herein, the field of soft robotic locomotion is still in its nascent phase. Numerous significant questions and concerns are yet to be addressed. In this section, we summerize the unique set of challenges related to their design, fabrication, modeling, and control.

\subsection{Design Challenges}

One of the primary challenges in designing soft mobile robots is creating a structure that is both flexible and strong enough to withstand the forces required for locomotion. When the actuators (i.e., limbs or body) are more compliant, they do not have the necessary strength. When they become stiff, compliance is lost. Hence, soft limbs and the body should be designed by embedding characteristics such as on-demand stiffness regulation without betraying its compliance \cite{arachchige2022hybrid,arachchige2021novel,amaya2021evaluation}. Mostly, current soft-limbed robot designs lack limb deformation range, control, strength, and gait stability. Hence, soft limbs must be designed with embedding adequate workspace, control, and bending stiffness while preserving inherent compliance. Moreover, soft robot assembly and gait generation methods should address limitations in stability during locomotion. The design must take into account the material properties, such as elasticity, viscosity, and compliance, to create a robot that can move in different environments. Additionally, designing the soft robot's shape, size, and texture to match the desired application is crucial.

\subsection{Fabrication Challenges}

Fabricating soft mobile robots presents several challenges due to the complex and deformable nature of these systems. They require materials that are highly compliant and deformable, yet also durable and resistant to wear and tear \cite{elango2015review}. Selecting the right materials for a given application can be challenging, as it requires balancing conflicting requirements such as stiffness, toughness, and biocompatibility. Further, soft mobile robots often require complex and specialized manufacturing techniques, such as 3D printing, molding, or casting, which can be difficult and expensive to implement \cite{joyee2019multi}. Moreover, these techniques often demand precise control of temperature, pressure, and humidity, which can be challenging to achieve. Additionally, soft mobile robots often require the integration of sensors, actuators, and control electronics into a compact and lightweight system. This can be challenging, as it requires careful selection of components and materials that can operate reliably and efficiently in the given environment. 

To address these fabrication challenges, we should develop new manufacturing techniques and materials that can be used to fabricate soft actuators more easily and reliably. These include techniques such as 3D printing, laser cutting, and soft lithography, as well as materials such as elastomers, hydrogels, and shape-memory polymers. Additionally, we should explore the use of modular design principles, which can enable the rapid prototyping and assembly of soft mobile robots using standardized building blocks. Alternatively, it aims to simplify robot construction and improve reliability by focusing on designing and developing simpler soft robotic units. Finally, developing efficient and scalable manufacturing processes that can produce soft mobile robots at a low cost is an important area of research, with potential applications in healthcare, entertainment, and education.

\subsection{Modeling Challenges}
Modeling soft mobile robots is a challenging task due to the complex behavior of the actuators they are made of. Soft actuators (i.e., body and limbs) are highly nonlinear systems that are difficult to model accurately due to the complex relationships between the inputs, outputs, and internal states of the robot \cite{godage2011dynamics}. These nonlinearities can arise from the material properties of the soft components, as well as the kinematics and dynamics of the robot. Soft actuators often involve multiple physical domains, such as mechanics, fluid dynamics, and electromagnetics, which interact in complex ways. This makes it challenging to develop models that capture all the relevant physical phenomena and their interactions. They exhibit significant variability in their behavior due to the inherent variability in their material properties and fabrication processes. This variability can make it difficult to develop accurate and robust models that can generalize across different instances of the same robot.

Modeling often requires the use of complex numerical simulations that can be computationally expensive and time-consuming. This makes it difficult to perform real-time control and optimization of the robot. Morever, validating the accuracy of robot models is challenging due to the lack of reliable experimental data that can be used for comparison. This makes it difficult to assess the performance of the models and identify areas for improvement. It is essential to accurately model the deformation of the soft robot's structure under various loads and constraints. Additionally, modeling the interaction between the robot and the environment, such as contact forces, friction, and deformation of the soft material, is crucial to understanding the robot's behavior and locomotion capabilities.

Addressing these modeling challenges requires the development of new modeling techniques that can capture the complex and dynamic behavior of the entire robot. This includes the use of advanced numerical simulations, machine learning algorithms, and experimental validation techniques. Additionally, there is a need for improved understanding of the underlying physical phenomena and material properties of soft robots, which can inform the development of more accurate and robust models.

\subsection{Control Challenges} 

Soft mobile robots present a unique set of control challenges that need to be addressed in order to achieve effective and reliable locomotion. Unlike traditional rigid robots, soft mobile robots are highly compliant and deformable, which makes them difficult to control using conventional methods. Soft actuators are highly nonlinear systems that exhibit complex interactions between their shape, motion, and external environment. This makes it difficult to develop accurate models and control strategies that can capture the full range of behaviors exhibited by these robots.

Soft mobile robots typically have a large number of degrees of freedom, which makes their state space high-dimensional. This can make it challenging to develop efficient control strategies that can operate in real-time and handle the complex interactions between different parts of the robot. Further, unconventional actuation mechanisms, such as pneumatics, SMAs, etc. can be difficult to control accurately and reliably. These systems can also be sensitive to noise and disturbances, which can make it challenging to maintain stable and predictable motion. Additionally, they often require large amounts of energy to achieve locomotion, which can limit their range and endurance. Developing energy-efficient control strategies is therefore an important challenge for soft mobile robots.

To address these challenges, researchers should develop new control strategies that can handle the nonlinear and high-dimensional nature of soft mobile robots. These include model-based control methods that use accurate models of the robot's dynamics and kinematics, as well as data-driven control methods that use machine learning algorithms to learn control policies from data. Additionally, it is required to explore novel actuation mechanisms, such as hybrid systems combining pneumatics, SMA and electromechanical techniques, which can provide more reliable and efficient actuation for soft mobile robots.

	\section{Conclusions\label{sec:Conclusion}}

    In this review, we provided a broad overview of recent research on wheelless terrestrial soft mobile robots. First, we classified various locomotion methods of soft-limbed and soft-bodied robots. Next, we detailed actuation mechanisms and modeling approaches thereof. The locomotion trajectory generation principles and related control systems were presented lately. Research challenges and mitigating approaches were discussed in the end. This article serves as a thorough resource manual for scholars engaged in the field of soft mobile robots focusing on wheelless terrestrial locomotion systems.

    Overall, the research challenges in soft mobile robots are multi-faceted and require interdisciplinary collaboration between materials science, engineering, and computer science to develop new solutions. Successful soft mobile robot designs require an integrated approach that considers the challenges of design, fabrication, modeling, and control. Despite their many challenges, researchers have made significant progress in developing new materials, manufacturing techniques, and design methods that can enable the development of soft mobile robots with unprecedented capabilities. As the field continues to evolve, it is likely that we will see innovative and transformative applications of soft mobile robots in the years to come. 
	
	\bibliographystyle{IEEEtran}
	\bibliography{refs}
	
	\ifCLASSOPTIONcaptionsoff
	\newpage
	\fi

 \ifCLASSOPTIONcaptionsoff
    \newpage
    \fi

\end{document}